\pdfoutput=1
\documentclass[11pt]{article}

\usepackage{emnlp2021}

\usepackage{times}
\usepackage{latexsym}
\usepackage{enumitem}
\usepackage{graphicx}
\usepackage{amsmath}
\usepackage{amssymb}
\usepackage{diagbox}
\usepackage{multirow}
\usepackage{subcaption}
\usepackage[export]{adjustbox}
\usepackage[many]{tcolorbox}
\usepackage[T1]{fontenc}
\usepackage[utf8]{inputenc}
 \usepackage{microtype}
\newcommand\mucovqa{\texttt{MuCo-VQA}}
\newcommand\tblock{\texttt{Transformer-Block}}
\newcommand\crossatt{\texttt{X-Att}}
\newcommand\languagevision{{language-and-vision}}
\hypersetup{final}

\NewTColorBox{NewBox}{ s O{!htbp} }{%
  floatplacement={#2},
  IfBooleanTF={#1}{float*,width=\textwidth}{float},
  colframe=blue!10!black,
  boxsep=0.1ex,
    left=2pt,
    right=2pt,
  }

\title{Towards Developing a Multilingual and Code-Mixed Visual Question Answering System by Knowledge Distillation}

\author{
Humair Raj Khan\thanks{$^*$These authors contributed equally to this work.}, \hspace{0.1cm} Deepak Gupta\footnotemark[1], \hspace{0.1cm} Asif Ekbal  \\
   Department of Computer Science and Engineering\\
   Indian Institute of Technology Patna, India  \\
  {\tt 
  khumairraj@gmail.com}, 
    {\tt\{deepak.pcs16,asif\}@iitp.ac.in}\\
  } 

\begin{document}
\maketitle
\begin{abstract}
Pre-trained language-vision models have shown remarkable performance on the visual question answering (VQA) task. However, most of the pre-trained models are trained by only considering monolingual learning, especially the resource-rich language like English. Training such models for multilingual setups demand high computing resources and multilingual language-vision dataset which hinders their application in practice. To alleviate these challenges, we propose a knowledge distillation approach to extend an English language-vision model (teacher) into an equally effective multilingual and code-mixed model (student). Different from the existing knowledge distillation methods, which only use the output from the last layer of the teacher network for distillation, our student model learns and imitates the teacher from multiple intermediate layers (language and vision encoders) with appropriately design distillation objectives for incremental knowledge extraction. We also create the large-scale multilingual and code-mixed VQA dataset in eleven different language setups considering the multiple Indian and European languages. Experimental results and in-depth analysis show the effectiveness of the proposed VQA model over the pre-trained language-vision models on eleven diverse language setups. 
% The datasets and source code are available here: \url{https://github.com/khumairraj/knowledge-distillation-MuCo-VQA}.

\end{abstract}

\begin{figure*}
    \centering
    \includegraphics[ width=0.98\textwidth]{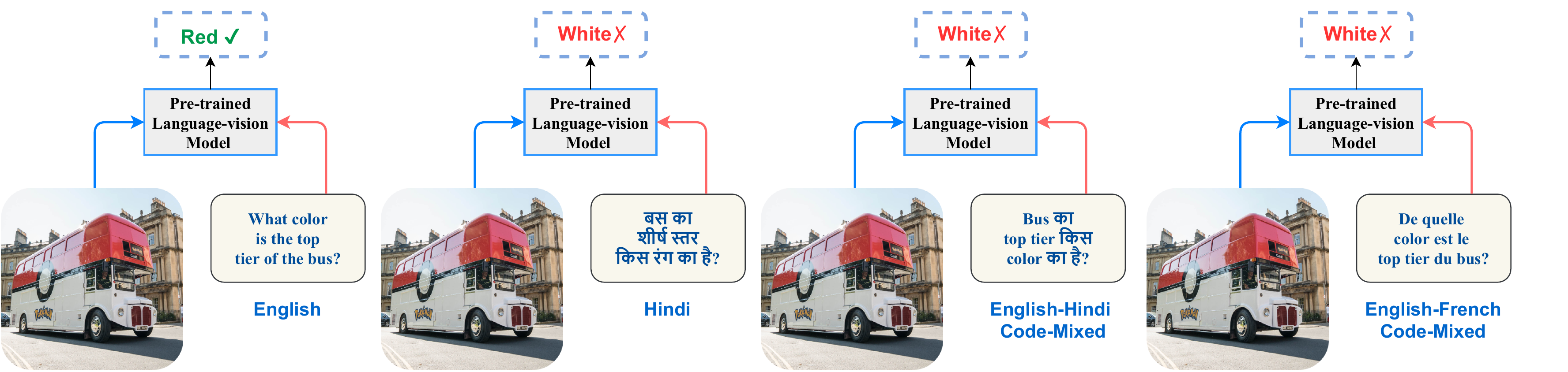}
    \caption{Example multilingual and code-mixed questions (same question), where pre-trained \languagevision{} models fails to correctly predict the answer except for English question.}
    \label{fig:examples}
\end{figure*}

\section{Introduction}
Visual Question Answering (VQA) is a challenging problem in computer vision (CV) and natural language processing (NLP) that have gained popularity due to its many-fold benefits ranging from assisting visually impaired users to establishing effective communication with robots \textit{via} intuitive interfaces.\\
%, interacting the visual data via natural language interfaces, and
% Solving the VQA task impacts on real-world applications such as assisting visually impaired users in gaining a better understanding of their physical and online environment, understanding
% the visual data via natural language interfaces, or using more efficient and intuitive interfaces to communicate with robots.
% Given a natural language question and an image, a VQA system requires reasoning over language-and-vision to answer the question. The language-and-vision reasoning requires the understanding of visual concepts, language semantics, alignment and relationships between the language and vision modalities. 
\indent The existing works \citep{tan-bansal-2019-lxmert,agrawal2017vqa} on VQA are mainly limited to the English questions, making it challenging to acknowledge progress in foreign languages. Moreover, the current language-vision models do not serve the purpose in the code-mixed setting, where the morphemes, words, phrases of one language are embedded into the other language. Since code-mixing has been a mean of communication in a multi-cultural and multi-lingual society, the next generation of artificial intelligence (AI) agents should be capable to understand the \textbf{M}ultilingual and \textbf{C}ode-\textbf{M}ixed (MCM) questions about the image.\\
\indent In the recent past, the pre-trained \languagevision{} models \cite{Su2020VL-BERT,tan-bansal-2019-lxmert,li2019visualbert,lu2019vilbert} have become the state-of the-arts for solving a variety of CV and NLP problems.
% such as image captioning, visual question answering, image retrieval, etc. 
However, the majority of these models are predominantly built for resource-rich languages like English. 
Therefore, their abilities to process and answer the MCM questions are limited (\textit{c.f.} Fig. \ref{fig:examples}).\\ 
% To build such pre-trained models which can understand the multilingual and code-mixed inputs requires the multilingual text-image pair datasets which are not readily available.\\ 
\indent To address this, we propose a highly effective and unified VQA method that allows us to extend the existing monolingual \languagevision{} models to multilingual (in \textbf{$6$ different languages}) and code-mixed (in \textbf{$5$ different code-mixed languages}) scenarios. Specifically, we develop a novel knowledge distillation \cite{hinton2015distilling} approach to distill the knowledge from the monolingual \languagevision{} transformer network (teacher model) to multilingual and code-mixed \languagevision{} transformer network (student model). This enables the student model to adapt to any language and code-mixed scenarios.\\
% in a multilingual and code-mixed setup
% allows the large scale \languagevision{} transformer model to learn ... 
% that enable the model to adapt on any languages and code-mixed scenarios. 
\indent In order to effectively transfer the knowledge from the teacher network to the student network, we introduce multiple distillation objectives which ensure the incremental knowledge extraction from multiple intermediate layers of \languagevision{} transformer model. These objectives are formulated to guide the student model to learn two key characteristics: \textbf{(1)} unified question representation across different languages, and
% learning the multilingual question embedding space such that vector spaces are aligned across the questions of different languages, and
\textbf{(2)} effective cross-modal (image-question) representation where the key objects in the image are attended irrespective of the language of the questions.
% i.e., same or near similar questions in different languages and code-mixed languages are closed in the vector space.\\
With these characteristics, we are able to build a unified VQA model, which can correctly predict the answers to multilingual and code-mixed questions.\\
\indent Furthermore, to combat the data scarcity, we also create a large-scale ($3.7$M image-question pairs) multilingual and code-mixed VQA dataset. Towards this, we utilize the English VQA1.0 \cite{agrawal2017vqa} dataset and extend it to multiple languages. We also create their code-mixing counterparts by designing the linguistically informed strategy to formulate the code-mixed question by mixing the words/phrases from the English question and the foreign language question. We evaluate our proposed approach on the created dataset and achieve $11.74\%$ average improvement across all the languages over the pre-trained language-vision model.\\
\textbf{Contributions:}
% \vspace{-0.5 em}
\begin{enumerate}[nolistsep]
 \item We devise a robust {knowledge distillation framework for multilingual and code-mixed VQA} by introducing multiple task-specific objective functions, which distill knowledge from the English pre-trained language-vision model to train and develop equally effective multilingual and code-mixed VQA system.
%  We introduce multiple task-specific objectives to effectively transfer the knowledge from teacher to student model. \\
%   To the best of our knowledge, this is the very first step where we attempt to propose a generic VQA method that produces the correct answer for the questions from multiple language pairs.\\
 \item We create the {large-scale ($3.7$M) multilingual and code-mixed VQA datasets} in
%  by following the linguistically-informed strategy to mix the words/phrases from two language pairs. It spans over 
 multiple languages: Hindi (\textit{hi}), Bengali (\textit{bn}), Spanish (\textit{es}), German (\textit{de}), French (\textit{fr}) and code-mixed language pairs: \textit{en-hi}, \textit{en-bn}, \textit{en-fr}, \textit{en-de} and \textit{en-es}. This dataset is publicly available here\footnote{\url{https://www.iitp.ac.in/~ai-nlp-ml/resources.html}}.
%  These datasets will be useful resources for the language and vision communities to develop multilingual and code-mixed language-vision downstream applications. \\
\item We demonstrate the effectiveness of our proposed single student model that can correctly predict the answers to the questions of the various language combinations (on {eleven ($11$) different language-vision setups}) including code-mixed setups over state-of-the-art pre-trained language-vision models.
% We demonstrate with detailed empirical evaluations (on {eleven ($11$) different language-vision setups}) and provided the in-depth analysis, which shows the effectiveness of our proposed approach. %our synthetic and neural generation technique 
\end{enumerate}

\section{Related Work}
% In this section, we present a summary of the related work on multilingual code-mixing, VQA datasets, and existing models on VQA. 
\paragraph{\textbf{Multilingual and Code-Mixing:}}
There is a recent trend in developing methods and resources for various NLP applications involving multilingual and code-mixed languages.  
% Towards this, \citet{garcia2015yet} introduces methods for solving various NLP problems with the inputs in English and Portuguese languages. 
%  \citet{al-rfou-etal-2013-polyglot,agerri2014ixa} created resources and tools for serving input from multiple languages.
Some of the works include question-answering \citep{raghavi2015answer,gupta-etal-2018-uncovering}, word embedding \cite{chen-cardie:2018:EMNLP,lample2018,word-embedding-for-code-mixed}, code-mixed text generation \cite{pratapa2018language,gonen2018language,gupta-etal-2020-semi}, code-mixed language modelling \cite{winata2018code,gonen2018language}, and other NLP tasks \cite{GUPTA18.486,gupta2016opinion,gupta2016hybrid,gupta2017smpost}. 
% There has been an attempt to create a VQA dataset and model from \citet{gupta-etal-2020-unified} but that focused on only two languages, \textit{viz.} English and Hindi. In contrast, this work introduces the datasets and method for VQA on a total of six different languages and eleven code-mixed settings.
\paragraph{\textbf{Visual Question Answering:}}
In the literature, various VQA datasets \cite{Silberman:ECCV12,gao2015you,agrawal2017vqa, goyal2017making} have been created to encourage multi-disciplinary research. The popular frameworks for VQA explore attention mechanisms to learn the joint representation of image and question \cite{mcb,mlb,yu2017multi,Kim2018BilinearAN}.
% Recently, with the success of Transformer \citep{vaswani2017attention} based pre-trained language models, multiple methods accounting language and vision for VQA have been proposed.
Recently, with the success of Transformer \citep{vaswani2017attention}, \citet{tan-bansal-2019-lxmert} proposed cross-modality framework, \textsc{LXMERT}, for learning the connection between vision and language. 
% \citet{li2019visualbert} introduced a pre-trained model, \textsc{VisualBERT} for joint language and vision representation. To learn the associations between the language and text, \textsc{VisualBERT} is pre-trained with two visually-grounded language model tasks on COCO image caption data \citep{chen2015microsoft}.
% \citet{Su2020VL-BERT} proposed a \textsc{VL-BERT} pre-trained model, which is based on single cross-modal architecture for language-vision task. The network is pre-trained with the masked language model objective and masked the visual feature classification. 
There are other notable works \citep{Su2020VL-BERT,zhou2020unified,li2020oscar}, where the Transformer-based models are pre-trained to learn the joint language-vision representation.
% and later used for the downstream tasks, such as VQA, image captioning, etc. 
Knowledge distillation has also been used in the literature for the VQA task for the optimal training strategy \cite{mun2018learning}, knowledge transfer from tri-modal to bi-modal \cite{do2019compact}, and the missing modalities \cite{cho2021dealing}. Unlike these, our current work focuses on knowledge transfer from the monolingual pre-trained language-vision model to the multilingual and code-mixed VQA. 
\section{Multilingual and Code-Mixed VQA Dataset}
\label{multilingual-dataset}
\paragraph{\textbf{Dataset Creation:}}
We follow the large-scale VQA dataset (VQAv1.0) released by \citet{agrawal2017vqa}.
The VQAv1.0 dataset contains the triplet information in the form of the question, image, and answers. \citet{gupta-etal-2020-unified}
introduced a VQA dataset named MCVQA which comprises of questions in Hindi and Hinglish (i.e. code-mixed English and Hindi). However, the approach to create MCVQA dataset has two major shortcomings: \textbf{(1)} algorithm is not scalable to other languages, and \textbf{(2)} it requires the NLP components (part-of-speech tagger, named entity recognizer,  transliteration, etc.) for the resource-scarce languages, which, themselves are an active research area for the resource-scare languages. 

% \textbf{(1).} the code-mixed questions generation algorithm is focused on the linguistic structure of the Hindi questions, which can not be scaled to the other language because the different word order typology, and

% \textbf{(2).} it uses the pre-trained Statistical Machine Translation (SMT) model to generate the candidate lexical translation,
% \footnote{For the code-mixed question generation, the Hindi words having the PoS tags (common noun, proper noun, spatio-temporal noun, adjective) and NE tags (\textit{LOCATION} and \textit{ORGANIZATION})
% are replaced with their best lexical translation.}, 
% which again introduce errors that 
% propagate to the code-mixed question generation pipeline; \textbf{(3)} the algorithm takes the foreign language question (Hindi) as input and it requires the PoS, NE taggers and transliteration model for the foreign language to generate the code-mixed questions. The quality of code-mixed question generation is highly dependent on the performance of these NLP components, which, themselves are a active research area for the resource-scare languages.  

To address these shortcomings, in this work, we create the large-scale ``\textbf{Mu}ltilingual and \textbf{Co}de-mixed \textbf{V}isual \textbf{Q}uestion \textbf{A}nswering'' (\mucovqa{}) dataset which supports the five ($5$) languages (\textit{hi}, \textit{bn}, \textit{es}, \textit{de}, and \textit{fr}) and five ($5$) different code-mixed settings (\textit{en-hi}, \textit{en-bn}, \textit{en-es}, \textit{en-de}, and \textit{en-fr}). To generate the code-mixed questions, we follow the matrix language frame (MLF) theory \cite{myers1997duelling} of code-mixed text.
% and adapt the technique from \citet{gupta-etal-2020-semi}.
According to MLF, a code-mixed sentence will have a dominant language (matrix-language) and inserted language (embedded-language). We utilize the Google machine translation to translate the English questions from VQAv1.0 dataset to the foreign language \textit{xx} $\in$ \{\textit{hi}, \textit{bn}, \textit{es}, \textit{de}, \textit{fr}\}. From the parallel questions (\textit{en-xx}), we learn the alignment of English words in the foreign language question. Given a pair of questions from the two languages, we identify the words following \citet{gupta-etal-2020-semi} from the English question and substitute their aligned counterparts (in foreign language question) with the identified English words to synthesize the English embedded code-mixed questions. 
% We show the samples from the created \mucovqa{} dataset in Figure \ref{fig:sample-mucovqa}.
Please see \textbf{Appendix} for the implementation details and samples of the \mucovqa{} dataset.
\paragraph{\textbf{Analysis:}}
Similar to the VQAv1.0 dataset, our created \mucovqa{} dataset consists of $248,349$ training and $121,512$ test questions for each of the five different languages and five code-mixed settings. 
We perform a qualitative analysis of this dataset by randomly selecting $5,00$ questions, each from \textit{en}, \textit{hi} and corresponding \textit{en-hi}. % from \mucovqa{} dataset. 
We seek annotation help from two bilingual (\textit{en}, \textit{hi}) experts to manually translate and create the code-mixed questions. Towards this, we compute the BLEU \citep{papineni2002bleu}, ROUGE \citep{lin2004rouge} and Translation Error Rate (TER) \citep{snover2006study} considering the manually created code-mixed questions as the gold standard; and the generated code-mixed questions from \mucovqa{} as the candidates. We compute the mean values of the individual scores obtained from both the experts. We found the BLEU: $78.34$, ROUGE-L: $91.13$, and TER: $8.23$, which show the generated code-mixed questions are close to the human formulated code-mixed questions.
% in Table \ref{cm-data-statistics}. Table \ref{cm-data-statistics} shows that the quality of the generated code-mixed questions are close to the human translated questions and generated code-mixed questions. 
The detailed analysis and statistics in terms of code-mixed complexity can be found in the \textbf{Appendix}.

\section{Methodology} \label{sec:methodology}
% \begin{figure*}[h]
% \centering
% % \input{final-model.pdf_tex}
% \caption{Architecture of the proposed multilingual VQA model. The input to the model is the multilingual question (one at a time). The bottom-right part of the image describes the \textit{Question Image Fusion} component.}
% \label{vqa-model}
% \end{figure*}
%This section describes the proposed methodology for multilingual and code-mixed VQA in detail.
% \paragraph{\textbf{Problem Statement}:} Given a multilingual or code-mixed question $\mathcal{Q}$ and image $\mathcal{I}$, the task is to correctly predict the answer $\hat{\mathcal{A}}$ from the answer vocabulary $\mathcal{A}$.
% % In this work, we are dealing with the questions from three different languages English, Hindi or English-Hindi code-mixed. 
% More formally:
% \begin{equation}
%     \hat{\mathcal{A}} = \argmax_{\hat{\mathcal{A}} \in \mathcal{A}} p(\hat{\mathcal{A}}|\mathbf{\mathcal{Q}}, \mathbf{\mathcal{I}}; \theta)
% \end{equation}
% where $\theta$ is the network parameters. 
% We depict the architecture of our proposed methodology in Figure \ref{vqa-model}.
Our proposed knowledge distillation framework for the VQA model is tailored to predict the answer for multilingual and code-mixed questions. 
We utilize LXMERT \cite{tan-bansal-2019-lxmert}, a pre-trained English vision-language model, as the teacher network to train our student network.
Our student network is inspired by the teacher network and has three components, \textit{viz.} \textbf{(1)} \textbf{MCM Question Encoder} that processes and effectively encodes the multilingual and code-mixed questions, \textbf{(2) Image Encoder} which learns the representation of the objects detected in the image, \textbf{(3) Cross-Modality Encoder}, that learns the joint feature representation by applying the cross-attention on the language and image features, and \textbf{(4) Answer Prediction}, which predicts the answer for MCM questions.
% Similar to \citet{tan-bansal-2019-lxmert}, we design the Transformer based hierarchical encoder to learn the representation of the image by exploiting the object representation obtained from  \cite{anderson2018bottom}. 
% The language and image encoder work in parallel to encode the respective modality, but to fuse both the modalities and learn the joint feature representation, we introduce the \textit{Cross-modality Encoder}, which learns the joint feature representation by applying the cross-attention on the language and image features.\\
% We utilize LXMERT, a pre-trained English vision-language model, as the teacher network to train our student network which predicts the answer for multilingual and code-mixed questions. 
%  (\textit{hi}, \textit{bn}, \textit{es}, \textit{de}, and \textit{fr}) and different code-mixed settings (\textit{en-hi}, \textit{en-bn}, \textit{en-es}, \textit{en-de}, and \textit{en-fr}). 
% % We choose LXMERT for two particular reasons. First, LXMERT has shown near state-of-the-art results on both language-vision understanding and vision reasoning tasks. Second, it outperforms (\textit{c.f.} Section \ref{sec:results}) the competitive pre-trained vision-understanding language model on \mucovqa{} dataset. 
% Our proposed approach for visual question answering is build upon the LXMERT model and consists of the neural components to tackle the multilingual and code-mixed questions.
\subsection{\textbf{Background}}
\label{lxmert-background}
\paragraph{Transformer Block: } For an input sequence $S^{l}=\{S_1^{l}, S_2^{l}, \ldots, S_{|S|}^{l}\}$ of length $|S|$ (which is the output of the $l^{th}$ transformer block) the $(l+1)^{th}$ transformer block computes the hidden states $S^{l+1}$ as follows: 
\begin{equation} \label{eq:transformer-block}
\begin{split}
 \hat{S}_i^{l+1} &= S_i^{l} + \text{MHA}(\text{LayerNorm}(S_i^{l}))\\
  {S}_i^{l+1} &= \hat{S}_i^{l+1} + \text{MLP}(\text{LayerNorm}(\hat{S}_i^{l+1}))\\
\end{split}
\end{equation}
where, $\text{MHA}(.)$ is Multi Head Attention \cite{vaswani2017attention}, $\text{LayerNorm}(.)$ is Layer Normalization \cite{ba2016layer} and $\text{MLP}(.)$ is a feed-forward network. 
Based on Eq. \ref{eq:transformer-block}, we define \texttt{Transformer-Block}(.) as a function of input $S^l \in \mathcal{R}^{|S| \times d}$ as follows:
\begin{equation} \label{eq:transformer-block-func}
\begin{split}
  {S}^{l+1} &=  \texttt{Transformer-Block}( {S}^{l})\\
\end{split}
\end{equation}
\subsection{Student Network} \label{sec:student-model} 
% Our proposed student network takes a question $\mathcal{Q}$, an image $\mathcal{I}$ as input and process them with the following components to predict the answer.
\paragraph{MCM Question Encoder:} The input question $\mathcal{Q}$ is tokenized using the WordPiece tokenizer~\cite{wu2016google} to form the sequence of tokens $\left\{t_1, t_2, \ldots, t_n\right\}$ with length $n$. We compute the word embedding $w_i$ for the $i^{th}$ token similar to the teacher network LXMERT.
% \begin{equation}
%         w_i = \mathrm{LayerNorm}\left({{TE}}({t}_i) + PE(i) + SE(\mathcal{Q}) \right) 
% \end{equation}
% where, ${{TE}}(.)$, $PE(.)$ and $SE(.)$ are the feed-forward networks which project the vectors for token, its position and input segment in the sequence. The addition of token, position and segment embeddings is subject to the Layer Normalization and the resultant vector $w_i \in \mathcal{R}^{d}$ is considered as word embedding.
Since, in this work, we deal with multilingual and code-mixed questions; therefore, we utilize Multilingual-BERT (M-BERT) \citep{devlin-etal-2019-bert} model as our language encoder. Multilingual-BERT is a single model pre-trained on the monolingual Wikipedia corpora from $104$ languages. The word embedding sequence $\left\{w_{0=\texttt{[CLS]}}, w_1, \ldots, w_n\right\}$ (with the \texttt{[CLS]} token) is passed to the stack of M-BERT encoders. Each M-BERT encoder consists of the MHA(.) layer followed by point-wise feed-forward network with the residual connection. We obtain the hidden state representation $H^M = \{h_0^{M},  h_1^{M}, \ldots, h_n^{M}\}$ from M-BERT having $M$ layers as follows:
\begin{equation} \label{eq:mbert}
\footnotesize
\begin{split}
 h_0^{1}, \ldots, h_n^{1} &= \text{M-BERT}^{l=1}(w_0, \ldots, w_n)\\
    h_0^{M}, \ldots, h_n^{M} &= \text{M-BERT}^{l=M}(h_0^{M-1}, \ldots, h_n^{M-1})
\end{split}
\end{equation}
For brevity, we will call $\{h_0^{M}, h_1^{M}, \ldots, h_n^{M}\}$ as $H= \{h_0,  h_1, \ldots, h_n\}$ in rest of the paper.

\paragraph{Image Encoder:} Given the input image $\mathcal{I}$, we extract $k$ objects $\{o_1, o_2, \ldots, o_k\}$ from \citet{anderson2018bottom}. For each object $o_j$, we obtain RoI features $r_j \in \mathcal{R}^{d_r}$ and bounding box co-ordinates $b_j \in \mathcal{R}^{d_b}$. We follow the object-relationship encoder from \citet{tan-bansal-2019-lxmert} to obtain the image representation. We first project RoI and co-ordinates via a feed-forward network to obtain $f_j$ and $p_j$, respectively. Then we obtain the object feature for the object $o_j$ as $u_j = (f_j + p_j)/{2}\in \mathcal{R}^{d}$ . 
% as follows:
% \begin{equation} \label{eq:image-features}
% \begin{split}
%  f_j &= \mathrm{LayerNorm}(\mathbf{W_f}r_j + c_f)\\
%  p_j &= \mathrm{LayerNorm} (\mathbf{W_p}b_j + c_p)\\
%  u_j &= {(f_j + p_j)}/{2}
% \end{split}
% \end{equation}
% where $\mathbf{W_f} \in \mathcal{R}^{d \times d_r}$ and $\mathbf{W_p} \in \mathcal{R}^{d \times d_b}$ are the weight matrices and $c_f \in \mathcal{R}^{d}$ and $c_p \in \mathcal{R}^{d}$ are the bias vectors. 
With $k$ objects in the image, we obtain the object feature matrix $U^0 \in \mathcal{R}^{k \times d}$. We employ the stack of \tblock{} (\textit{c.f.} Eq . \ref{eq:transformer-block-func}) to encode the image. For the first \tblock{}, we fed the object feature matrix $U^0$ and obtain
the hidden state representations $u_1^{1}, \ldots, u_k^{1}$. Subsequently, we obtain the final image representation $U= U^N \in \mathcal{R}^{k \times d}$ from the last layer ($N$) of \tblock{} as follows:
\begin{equation}
\footnotesize
u_1, \ldots, u_k =\tblock{}(U^{N-1})
\end{equation}
\paragraph{Cross-Modality Encoder:}
Given the MCM question representation $H \in \mathcal{R}^{n \times d}$ and image representation $U \in \mathcal{R}^{k \times d}$, similar to \citet{tan-bansal-2019-lxmert}, we aim to compute the cross-modal representations using the layers of \tblock{}. For a given layer $l$, the cross-modality encoder consists of two cross-attention layers (one from question to image another from image to question) and two \tblock{} for each modality. Cross-attention layer $\crossatt{}(.)$ takes the query vector $x^q$ of the representation $x$ from one of the modals and compute the attention weight $\alpha_j = softmax(x^q.y_j^k)$ with the key vectors= $y_j^k$ from the other modality. Thereafter, it computes the final cross-modal representation $\overline{x}= \sum \alpha_j y_j^v$ as the weighted average of the set of value vectors $\{y^v\}$. For the cross-modal representation $\overline{H}^l \in \mathcal{R}^{n \times d}$ from the $l^{th}$ layer of the question, we apply the \crossatt{} followed by the \tblock{} operation as follows:
\begin{equation} \label{eq:cross-modal-question}
\footnotesize
\begin{split}
  \widetilde{h}_i^{l}&= \crossatt{}(h_i^{l-1}, [u_1^{l-1}, u_2^{l-1}, \ldots, u_k^{l-1}] )\\
  \widetilde{H}^l & = [\widetilde{h}_0^{l}, \widetilde{h}_1^{l}, \ldots, \widetilde{h}_n^{l} ] \in \mathcal{R}^{n \times d}\\
 \overline{H}^l &= \tblock{}(\widetilde{H}^{l})
\end{split}
\end{equation}
Similarly, the cross-modal representation $ \overline{U}^l$ for the image considering the question as another modal is computed.
% is as follows:
% \begin{equation} \label{eq:cross-modal-image}
% \footnotesize
% \begin{split}
%  \overline{U}^l &= \tblock{}(\widetilde{U}^{l})
% \end{split}
% \end{equation}
We use the $L$ layers of cross-modal encoders to encode the cross-modal representation.
\paragraph{Answer Prediction: } To predict the answer for the multilingual question, we take the output of question from the last ($L^{th}$) cross-modal encoders. We use the \texttt{[CLS]} token representation $\overline{h}_{\texttt{[CLS]}}^{L} \in \mathcal{R}^{d}$ and predict the answer as follows:
\begin{equation} \label{eq:answer-predict}
\footnotesize
\begin{split}
  \overline{P} &= gelu(\mathbf{W_P}\overline{h}_{\texttt{[CLS]}}^{L} + c_{P})\\
  p({\mathcal{A}_{i}}|\mathbf{\mathcal{X}}; \theta^{S}) &= \sigma(\mathbf{W}_{i} \overline{P}+ c_{i})
\end{split}
\end{equation}
where, $\mathbf{W_P} \in \mathcal{R}^{2d \times d}$ is the weight matrix and $c_{P} \in \mathcal{R}^{2d}$ is the bias vector. $\sigma$ denotes the sigmoid function. $W_i$ and $c_i$ are the $i^{th}$ entry of weight matrix $W \in \mathcal{R}^{d \times |\mathcal{A}|}$ and bias vector $c \in \mathcal{R}^{ |\mathcal{A}|}$.
$\overline{h}_{\texttt{[CLS]}}^{L} \in \mathcal{R}^{d}$ is the hidden state representation of \texttt{[CLS]} token obtained from cross-modality encoder. $|\mathcal{A}|$ is the length of the answer vocabulary. $\mathcal{X}$ is the set of input $\{\mathcal{Q}, \mathcal{I}\}$. $p({\mathcal{A}_{i}}|\mathbf{\mathcal{X}}; \theta^{S})$ is the probability of the $i^{th}$ answer from answer vocabulary $\mathcal{A}$.
\subsection{Distillation Objectives}
In our knowledge-distillation framework, we propose multiple objectives to transfer the knowledge from the monolingual Teacher network (with $\theta^T$ parameters) to the MCM Student network (with $\theta^S$ parameters): % which are as follows:
\paragraph{Objective 1 - \texttt{CLS} Token Distillation:} The \texttt{[CLS]} token embedding learned at the cross-modality encoder represents the semantics of the monolingual question-image pair in teacher network and MCM question-image pair in student network. We argue that it should learn a similar representation to correctly predict the answer irrespective of the language. Towards this, we compute the \texttt{[CLS]} token loss by computing the Mean Squared Error (MSE) between the vector representation learned at the Cross-modality Encoder in the teacher network and student network.
\begin{equation} \label{eq:clas-loss}
\footnotesize
\begin{split}
\mathcal{L}_{CLS} = \sum_{i=1}^{i=|L|} \sum_{j=1}^{j=|MH|} \mathbf{MSE}(\overline{h}_{(i,j,[CLS])}^{T}, \overline{h}_{(i,j,[CLS])}^{S})
\end{split}
\end{equation}
where, $\overline{h}_{(i,j,[CLS])}^{T} \in \mathcal{R}^{d}$ and $\overline{h}_{(i,j,[CLS])}^{S} \in \mathcal{R}^{d}$ are the representation of the \texttt{[CLS]} token obtained from the $i^{th}$ cross-modal encoder layer under $j^{th}$ attention head from teacher and student network, respectively. $|MH|$ is the number of attention head in the \tblock{}.

\paragraph{Objective 2 - Object Attention Distillation:} The answer to a given question is defined by the object detected in the image. It is to be noted that the answer to a question is independent of the language. We argue that in order to correctly predict the answer to MCM questions, the student network should attend the same object as the teacher network. This helps in aligning the question representation across different languages to the object representation and thus assists towards learning the effective language-agnostic cross-modal representation of the question-image pair. Towards this, we compute the object attention loss ($\mathcal{L}_{object}$), which measures the MSE between the raw score vectors $z \in \mathcal{R}^{k}$ (obtained using the dot product between \texttt{[CLS]} token's query vector and set of object's key vector) learned at the Cross-modality Encoder in the teacher network and student network.
\begin{equation} \label{eq:object-loss}
\footnotesize
\begin{split}
\mathcal{L}_{object} = \sum_{i=1}^{i=|L|} \sum_{j=1}^{j=|MH|} \mathbf{MSE}(z_{(i,j)}^{T}, z_{(i,j)}^{S})
\end{split}
\end{equation}
where, $z_{(i,j)}^{T} \in \mathcal{R}^{k}$ and $z_{(i,j)}^{S} \in \mathcal{R}^{k}$ are the vector raw scores obtained from the $i^{th}$ layer under the $j^{th}$ attention head from Teacher and Student network, respectively.

\paragraph{Objective 3 - Prediction Distillation:}  In addition to imitating the behaviors of intermediate layers, we
also use the knowledge distillation to mimic the predictions of teacher network. Specifically, we penalize the binary cross-entropy loss between the answer probabilities obtained from the teacher and student network.
% \vspace{-1em}
\begin{equation} \label{eq:prediction}
\footnotesize
\begin{split}
\mathcal{L}_{pred} &= -\sum_{i=1}^{i=|\mathcal{A}|} p(\mathcal{A}_i|\mathcal{X}; \theta^T) log(p(\mathcal{A}_i|\mathcal{X}; \theta^S)) + \\
& (1- p(\mathcal{A}_i|\mathcal{X}; \theta^T)) log(1- p(\mathcal{A}_i|\mathcal{X}; \theta^S))
\end{split}
\end{equation}

% \begin{multline*}
%  \label{eq:prediction}
% \footnotesize
% \mathcal{L}_{pred} = -\sum_{i=1}^{i=|\mathcal{A}|} p(\mathcal{A}_i|\mathcal{X}; \theta^T) log(p(\mathcal{A}_i|\mathcal{X}; \theta^S)) + \\
% (1- p(\mathcal{A}_i|\mathcal{X}; \theta^T)) log(1- p(\mathcal{A}_i|\mathcal{X}; \theta^S))
% \end{multline*}

% \begin{equation} \label{eq:prediction}
% \footnotesize
% \begin{split}
% \mathcal{L}_{pred} = -\sum_{i=1}^{i=|\mathcal{A}|} p_{(\mathcal{A}_i, s ;)} log(p^{S}(\mathcal{A}_i|\mathcal{X}; \theta^S)) + \\
% (1- p^{T}(\mathcal{A}_i|\mathcal{X}; \theta^T)) log(1- p^{S}(\mathcal{A}_i|\mathcal{X}; \theta^S))
% \end{split}
% \end{equation}

\paragraph{Objective 4 - Negative Log-likelihood Loss:} 
% We also use the negative log-likelihood (NLL) loss to further train the student network to correctly predict the answer.
We also penalize the binary cross-entropy loss between the gold answer probability $y_i$ and model's predicted probability $p(\mathcal{A}_i|\mathcal{X}; \theta^S)$ obtained from the student network. 
% We define the loss function $\mathcal{L}_{nll}$ similar to the Eq. \ref{eq:prediction}, except the teacher network's prediction probability $p^T$(.) is replaced by the gold answer probability $y_i$.\\
\vspace{-1em}
\begin{equation} \label{eq:nll}
\footnotesize
\begin{split}
\mathcal{L}_{nll} &= -\sum_{i=1}^{i=|\mathcal{A}|} y_ilog(p(\mathcal{A}_i|\mathcal{X}; \theta^S)) +  \\
&(1-y_i)log(1- p(\mathcal{A}_i|\mathcal{X}; \theta^S))
\end{split}
\end{equation}
% \vspace{-2em}
% \subsection{Learning}
% To apply the knowledge distillation, first we need to
% train our teacher network. Similar to Eq. \ref{eq:nll}, we first train our teacher network having $\theta^{T}$ parameters with English question from VQAv1.0 dataset. Thereafter, the teacher network's parameters are frozen (and kept on evaluation mode to predict the answer for English question) and the student network is trained with \mucovqa{} dataset by minimizing the following objective function:
% \indent We train our knowledge distillation framework with the following loss:
% \begin{equation} \label{eq:student-loss}
% \footnotesize
% \begin{split}
% \mathcal{L} = \mathcal{L}_{\texttt{CLS}} + \mathcal{L}_{\texttt{object}} + \mathcal{L}_{\texttt{pred}} + \mathcal{L}_{\texttt{nll}}
% \end{split}
% \end{equation}

\subsection{Learning}
To apply the knowledge distillation, first we need to
train our Teacher network, % We first train our Teacher network 
having $\theta^{T}$ parameters with English question from the VQAv1.0 dataset. Thereafter, the Teacher network's parameters are frozen and the Student network is trained with the following objective function:
\begin{equation} \label{eq:student-loss}
\begin{split}
\mathcal{L} = \mathcal{L}_{\texttt{CLS}} + \mathcal{L}_{\texttt{object}} + \mathcal{L}_{\texttt{pred}} + \mathcal{L}_{\texttt{nll}}
\end{split}
\end{equation}
% During training, we fed the Teacher network with the English question and image, and the multilingual and code-mixed question (one language at a time) and their associated images are passed to the Student network.
During training, the Teacher network is fed with the English question and the corresponding image, and the Student network is fed with multilingual and code-mixed questions (one language at a time) and the corresponding image.

% During training, we fed the Teacher network with the English question and image, and the multilingual and code-mixed question (one language at a time) and their associated images are passed to the Student network.
\section{Dataset and Experiments} \label{sec:experiments}
\subsection{Datasets} We evaluate our proposed knowledge distillation framework on the \mucovqa{} dataset having eleven different language setups, and the MCVQA dataset \cite{gupta-etal-2020-unified} that consists of \textit{en}, \textit{hi}, \textit{en-hi} language setups. We train the Student network with the training dataset from all these languages. 
We take out $5\%$ of the training dataset as the validation dataset for evaluating and selecting the best Student model. The best Student model is used to evaluate the performance of the \mucovqa{} test dataset in all the language setups. For evaluation, we follow the accuracy metric as defined in \citet{agrawal2017vqa}.

\subsection{Implementation Details} 
We use the pre-trained Multilingual BERT\footnote{\url{https://github.com/google-research/bert/blob/master/multilingual.md}} having $12$ encoder layers, each having $12$ attention heads and a hidden dimension of $768$ for each token. In our proposed knowledge distillation framework, we train the model on \mucovqa{} training dataset for $16$ epochs. 
Since the input image remains the same for the LXMERT model and our proposed model, we initialize our image encoder weights with the LXMERT object relationship encoders' weights.

We set the maximum question length to $20$ words. The numbers of objects extracted from the image is $k=36$ and the dimension of bounding box coordinates and RoI features are $d_{b}=4$ and $d_r=2048$, respectively. For the Teacher network, the language encoder has $M=9$ layers, the image encoder has $N=5$ layers and the cross-modality encoder has the $L=5$ layers. Similarly in the Student network, the values of these layers are $M=12, N=5, L=5$. During training, we fine-tune the top $4$ M-BERT encoders and the top $2$ image encoders. We learn the cross attention layer from scratch to align the multilingual and vision embeddings. For the \texttt{CLS} token distillation, we set the layers $i \in \{1, 4\}$ and attention head $j \in \{1, 4, 5\}$.
Optimal values of the hyperparameters are chosen based on the model performance on the development set of \mucovqa{} dataset.

\subsection{Baselines} \label{sec:baselines}
We compare the performance of the proposed network with the following baseline models.\\
\textbf{(1)} \textbf{LXMERT}: We train the individual LXMERT model on the training dataset of each language from \mucovqa{} dataset and evaluate the performance on the respective test dataset. \\
\textbf{(2)} \textbf{Joint LXMERT}: We train the single LXMERT model on all the training datasets of each language from \mucovqa{} dataset and evaluate the performance on the respective test dataset. \\
\textbf{(3)} \textbf{Joint LXMERT+ M-BERT}: This baseline is similar to the Joint LXMERT but the monolingual language encoder is replaced with a multilingual M-BERT encoder. \\
\textbf{(4)} \textbf{VL-BERT} \cite{Su2020VL-BERT}: 
We also compare the performance of our proposed model with the VL-BERT base model (\texttt{vl-bert-base-e2e.model}). We train separate VL-BERT model on the training dataset of each language from the \mucovqa{} dataset and evaluate the performance on the respective test dataset. \\
\textbf{(5)} \textbf{VisualBERT} \cite{li2019visualbert}: 
Similar to the LXMERT, we also compare the performance of our proposed network on the \mucovqa{} dataset with the VisualBERT monolingual model.

\begin{table*}[]
\resizebox{\textwidth}{!}{%
\begin{tabular}{l|c|cccccccccccccc}
\hline
 & \textbf{Models} & \textbf{\textit{bn}} & \textbf{\textit{en-bn}} & \textbf{\textit{de}} & \textbf{\textit{en-de}} & \textbf{\textit{es}} & \textbf{\textit{en-es}} & \textbf{\textit{fr}} & \textbf{\textit{en-fr}} & \textbf{\textit{hi}} & \textbf{\textit{en-hi}} & \textbf{\textit{en}} & \textbf{\textit{Average}} \\ \hline \hline
\multirow{6}{*}{\rotatebox[origin=c]{90}{\textbf{Monolingual}}} &
 \begin{tabular}[c]{@{}c@{}}LXMERT\\ \cite{tan-bansal-2019-lxmert}\end{tabular}  & 60.74 & 64.95 & 67.95 & 70.52 & 68.66 & \textbf{71.27} & 68.43 & 71.12 & 59.83 & 69.95 & \textbf{73.57} & 67.90 \\
&\begin{tabular}[c]{@{}c@{}}VL-BERT\\ \cite{Su2020VL-BERT}\end{tabular} & 58.53 & 61.30 & 64.29 & 65.33 & 64.79 & 66.09 & 64.72 & 65.84 & 59.40 & 65.28 & 67.30 & 63.89 \\ 
&\begin{tabular}[c]{@{}c@{}}VisualBERT\\ \cite{lu2019vilbert}\end{tabular} 
&61.45&	64.20 &	66.74&	67.49&	67.31&	67.42&	66.35&	67.21&	59.68&	63.67&	68.12&	65.42 \\ \hline

\multirow{8}{*}{\rotatebox[origin=c]{90}{\textbf{Multilingual}}} &\begin{tabular}[c]{@{}c@{}}Joint LXMERT\\ \cite{tan-bansal-2019-lxmert}\end{tabular} & 48.44 & 60.07 & 58.02 & 62.79 & 58.41 & 63.07 & 58.34 & 62.96 & 50.28 & 61.52 & 65.41 & 59.02 \\ 
&Joint LXMERT+ M-BERT & 55.68 & 56.89 & 57.73 & 58.01 & 57.87 & 58.34 & 57.45 & 57.82 & 56.22 & 57.18 & 58.78 & 57.45 
\\ \cline{2-14}
&\textbf{Proposed Approach} & {\textbf{69.62}} & \textbf{70.19} & \textbf{70.89} & \textbf{70.80} & \textbf{71.14} & 71.11 & \textbf{70.93} & \textbf{71.13} & \textbf{70.23} & \textbf{70.78} & 71.66 & \textbf{70.76} \\  \cline{2-14}
&\quad $-\mathcal{L}_{CLS}$ & 67.95 & 68.50 & 69.18 & 69.06 & 69.45 & 69.59 & 69.23 & 69.42 & 68.55 & 69.05 & 69.91 & 69.08 \\ 
&\quad $-\mathcal{L}_{object}$ & 66.02 & 66.56 & 67.30 & 67.32 & 67.60 & 67.74 & 67.32 & 67.56 & 66.66 & 67.13 & 68.80 & 67.27 \\ 
&\quad $-\mathcal{L}_{pred}$ & 68.17 & 68.77 & 69.40 & 69.32 & 69.65 & 69.88 & 69.42 & 69.62 & 68.77 & 69.30 & 70.11 & 69.31 \\ 
&\quad $-\mathcal{L}_{nll}$ & 68.17 & 68.83 & 69.50 & 69.41 & 69.70 & 69.88 & 69.48 & 69.47 & 68.80 & 69.35 & 70.28 & 69.35 \\ \hline
\end{tabular}%
}
\caption{Performance comparison between the state-of-the-art baselines and our proposed model on the \mucovqa{} dataset. All the numbers are shown in \% and denote the overall accuracy. 
% $\dagger$ denotes that the comparison between
% monolingual and proposed approach may not be fair since the former models is trained and evaluated separately with the respective language dataset and have the different model for each language setup while the later is a single multilingual unified model.
}
\label{tab:main-results}
\end{table*}

\begin{table}[h]
\resizebox{\linewidth}{!}{%
\begin{tabular}{ccccc}
\hline
\textbf{Models} & \textbf{\textit{en}} & \textbf{\textit{hi}} & \textbf{\textit{en-hi}} & \textbf{\textit{Average}} \\ \hline
LXMERT \cite{tan-bansal-2019-lxmert} & \textbf{73.02} & 63.33 & 68.77 & 68.37\\ 
VL-BERT \cite{Su2020VL-BERT} &67.28& 59.32 & 63.28 & 63.29  \\
VisualBERT \cite{li2019visualbert} &68.04& 59.69 & 63.62 & 63.78  \\
\citet{gupta-etal-2020-unified} & 65.37 & 64.51 & 64.69 &  64.85\\ \hline
\textbf{Proposed Approach} & 71.37 & \textbf{69.94 }& \textbf{69.47} & \textbf{70.26}\\ \hline
\end{tabular}%
}
\caption{Performance comparison of different models on the MCVQA dataset.}
\label{tab:mcvqa-results}
\end{table}

% \usepackage{graphicx}
\begin{table}[]
\resizebox{\linewidth}{!}{%
\begin{tabular}{ccccc}
\hline
\textbf{Language} & \textbf{Number} & \textbf{Other} & \textbf{Yes/No} & \textbf{Overall} \\ \hline \hline
\textit{\textbf{en}} & 51.15 & 64.56 & 88.02 &  71.66 \\ 
\textit{\textbf{bn}} & 50.62 & 62.16 & 85.97 &  69.62 \\ 
\textit{\textbf{en-bn}} & 50.78 & 62.89 & 86.45 & 70.19 \\
\textit{\textbf{de}} & 50.86 & 63.50 & 87.49 & 70.89 \\ 
\textit{\textbf{en-de}} & 50.84 & 63.34 & 87.45 &  70.80\\
\textit{\textbf{es}} & 50.93 & 63.95 & 87.54 & 71.14 \\ 
\textit{\textbf{en-es}} & 50.94 & 64.19 & 87.68 & 71.11 \\
\textit{\textbf{fr}} & 50.95 & 63.47 & 87.61 &70.93   \\ 
\textit{\textbf{en-fr}} & 51.01 & 63.84 & 87.58 & 71.13 \\
\textit{\textbf{hi}} & 50.72 & 62.57 & 87.01 &  70.23\\ 
\textit{\textbf{en-hi}} & 50.95 & 63.30 & 87.43 &  70.78\\ \hline \hline
\end{tabular}%
}
\caption{Performance of our proposed model on different answer types across all the language setups in \mucovqa{} dataset}
\label{tab:lang-results}
\end{table}

\subsection{Results} \label{sec:results}
We report the performance of the baseline models and our proposed model on \mucovqa{} dataset in Table \ref{tab:main-results}. We also reported the answer-type wise results on \mucovqa{} dataset in Table \ref{tab:lang-results}.
% Amongst all the baselines, LXMERT achieves an average overall accuracy of $67.90$ having $73.57$ accuracy on $en$ language, which is also the teacher network in our knowledge distillation framework. In multilingual baselines, Joint-LXMERT reported an average of $59.02$ overall accuracy with maximum of $65.41$ on \textit{en} language. Joint LXMERT+M-BERT model achieves slightly lower performance ($57.45$) compared to the Joint LXMERT because the question encoder learns from the M-BERT could not be aligned properly only with the cross-entropy loss.
Our proposed model achieves $70.76$ overall accuracy and outperforms the best monolingual and multilingual baselines with significant improvements of $2.86$ and $11.74$, respectively. Our proposed approach also outperforms the state-of-the-art model on MCVQA dataset (\textit{c.f.} Table \ref{tab:mcvqa-results}) with considerable performance improvement of $5.07\%$. 
We could not observe a similar improvement on \textit{en} language, because the LXMERT teacher model (\textit{en}) is already pre-trained with the English VQA dataset.
% Our single unified model achieves the overall accuracy of $71.66$ on \textit{en} language, which is slightly lower than that LXMERT teacher network performance of $73.57$. 

It is to be noted that each monolingual model is trained separately with the respective language dataset and has a different model for each language setup. 
The results conclude two important claims: \textbf{(1)} effectiveness of knowledge distillation approach to handle MCM questions, and \textbf{(2)} scalability of our proposed single unified VQA model that can deal with questions from all the languages and their code-mixed setups.\\
% \paragraph{Ablation Study} \label{sec:ablation}
\indent We also perform the ablation study (\textit{c.f.} Table \ref{tab:main-results}) on different distillation objective functions.
% train our proposed model by removing one loss function at a time. 
% The results are reported in Table \ref{tab:main-results}. 
The results show that Object Attention Distillation ($\mathcal{L}_{object}$) is the most contributing objective function, removal of which leads to the $3.49\%$ decrements in the overall average accuracy. We also observe the importance of the \texttt{CLS} Token Distillation ($\mathcal{L}_{\texttt{CLS}}$). This is the key loss function responsible to align the same multilingual and code-mixed questions in the vector space, and removing it leads to $1.68\%$ decrements in overall average accuracy. 
Similarly, we observe $1.45\%$ and $1.41\%$ performance drops after the removal of $\mathcal{L}_{pred}$ and $\mathcal{L}_{nll}$ objective functions, respectively. The observed improvements over the multilingual baselines are statistically significant as $p< 0.05$ for the t-test using \citet{statistical-test}.
Please see the \textbf{Appendix} for additional results. 
% Similarly, we observe $1.45\%$ and $1.41\%$ performance drops after the removal of $\mathcal{L}_{pred}$ and $\mathcal{L}_{nll}$ objective functions, respectively.
 \begin{figure}[t]
    \centering
    \includegraphics[ width=\linewidth]{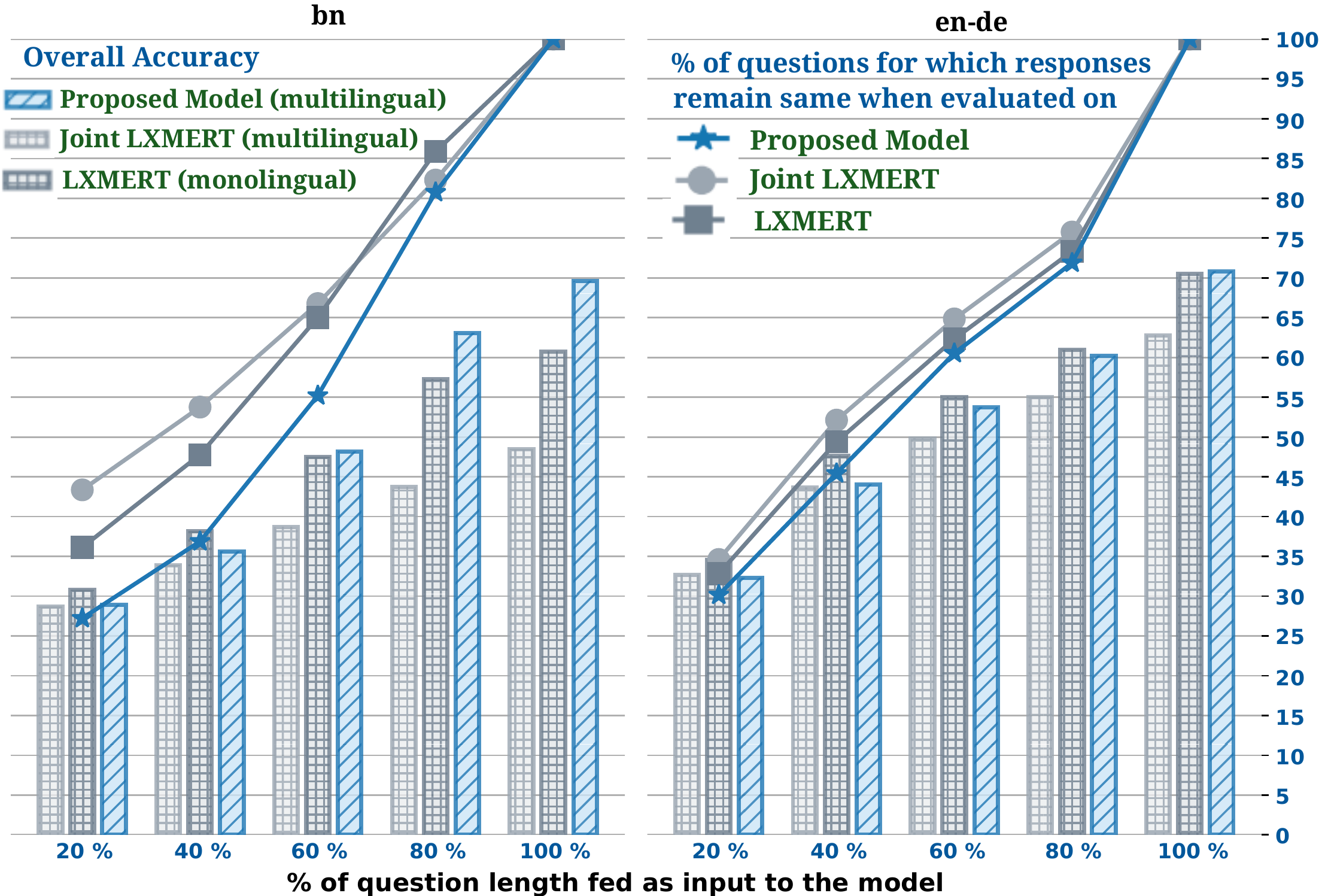}
    \caption{Performance comparison of different models for question understanding by varying the partial question as input to the model.}
%X-axis shows length of partial question (in %)
%fed as input to the state-of-the-art (LXMERT and Joint LXMERT) models and the proposed model. Y-axis shows %percentage of questions for which responses of these %partial questions are the same as full questions(line %chart) and VQA accuracy of corresponding model for the %partial questions(bar chart). This comparison shows that %the proposed model does not jump to quick conclusions by %looking at partial questions as both the metrics are low %for the proposed model for incomplete questions. At full %length questions, these metric however are high for the %proposed model indicating that the model is sensitive %enough to questions in different language.
    \label{fig:question-understanding-incomplete-main}
\end{figure}
\begin{figure}[]
  \subfloat[\centering \textbf{Joint LXMERT}]
  {{\includegraphics[width=0.45\linewidth]{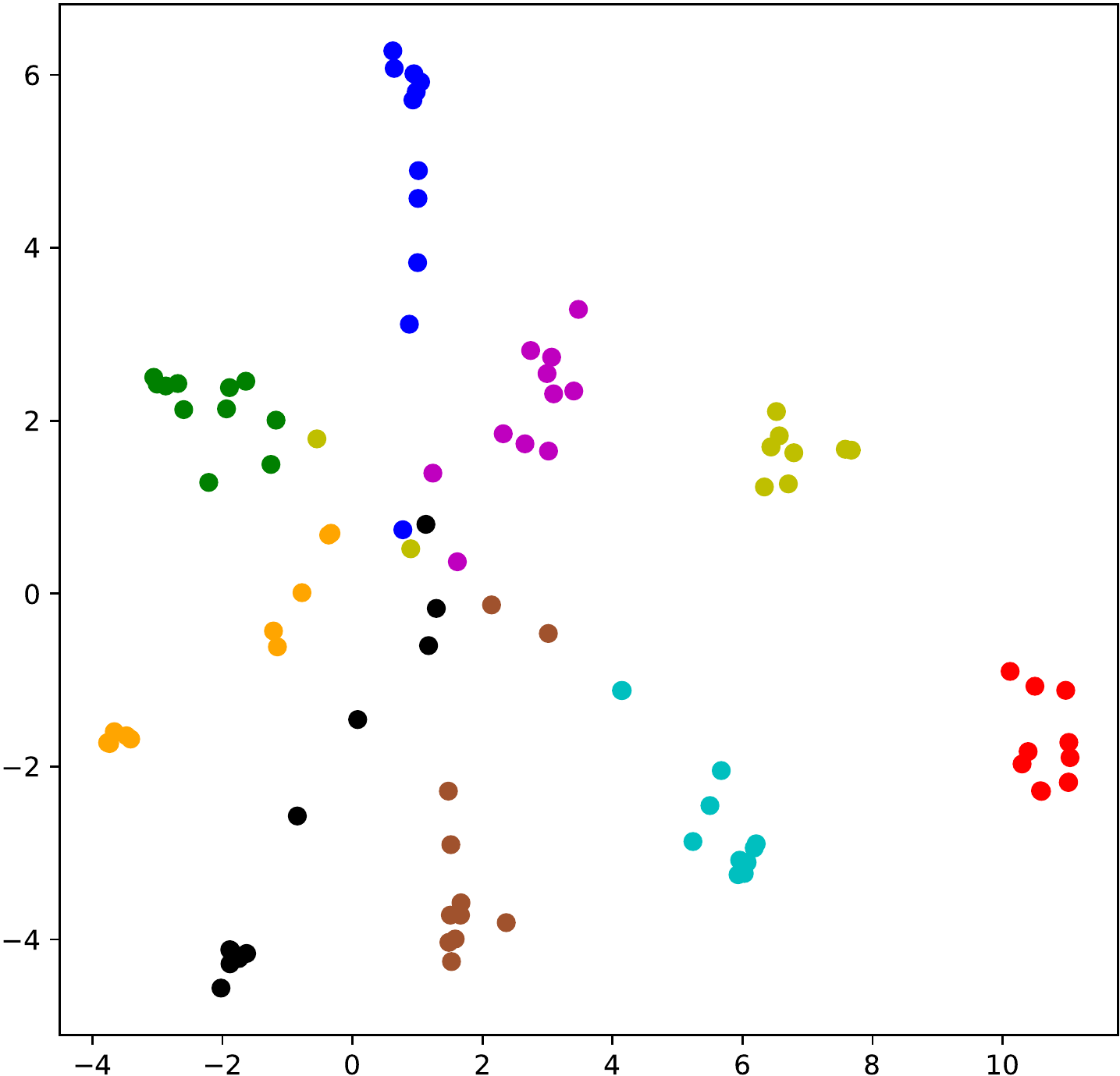} }}%
    \quad
    \subfloat[\centering \textbf{Proposed Approach}]
    {{\includegraphics[width=0.45\linewidth]{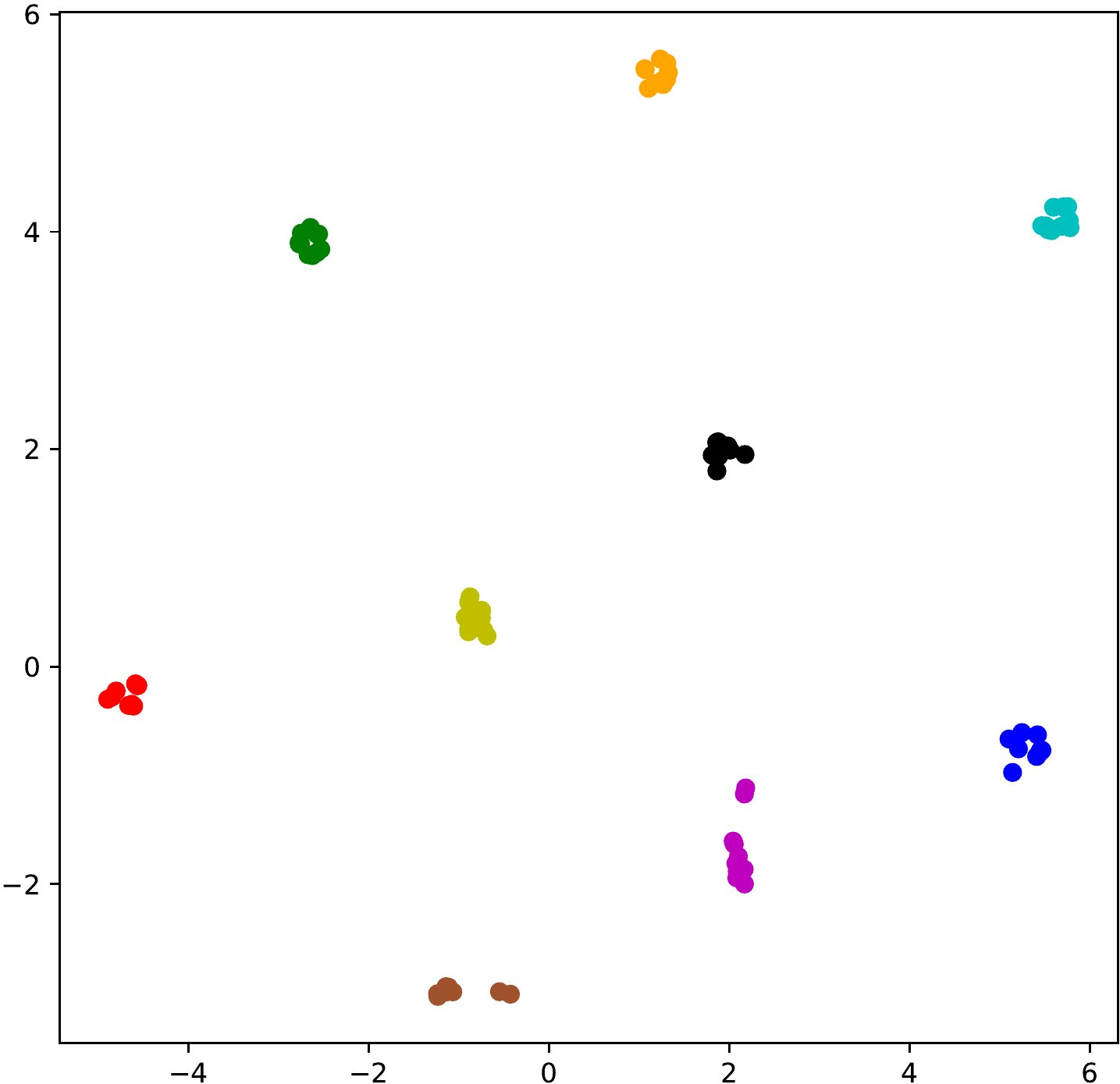} }}%
     \vspace{5pt}
\captionof{figure}{t-SNE visualization for MCM questions in all eleven language setups. For proposed approach (b), we observe that the question representations of the same questions (shown in the same color) in different languages are very close in vector space unlike the Joint LXMERT model (a).}
\label{question-reps}
 \end{figure}

% \vspace{10mm}

\begin{figure}[t]
 \begin{minipage}{.1\textwidth}
  \centering
 {{\includegraphics[width=\linewidth]{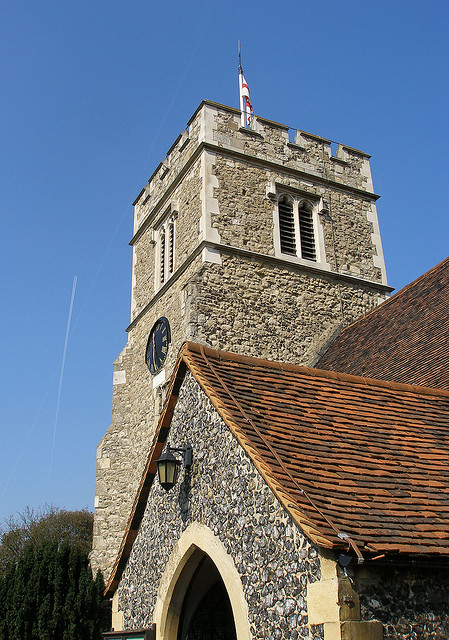} }}%
 \end{minipage}
    \quad
    \begin{minipage}{.35\textwidth}
  {{\includegraphics[width=\linewidth]{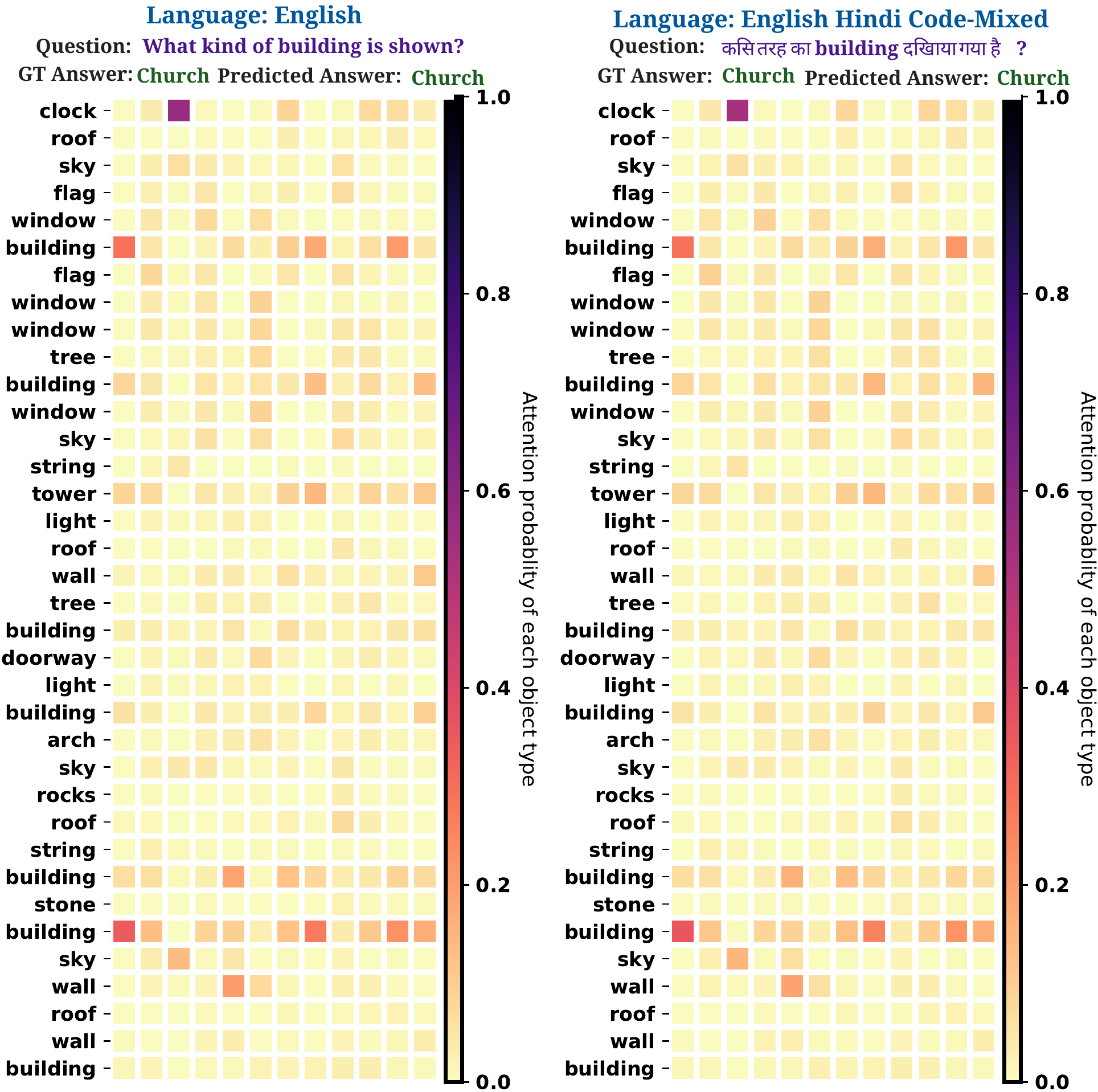}}}%
  \end{minipage}
\captionof{figure}{Heatmap of the learned attention weight for objects in the image from our proposed model. The proposed model is able to focus on the same object and correctly predict the answer irrespective of the language of the question. x-axis shows the heads of self-attention.}
\label{fig:question-object-alignment-main}
 \end{figure}

 \begin{figure}[t]
    \centering
    \includegraphics[ width=\linewidth]{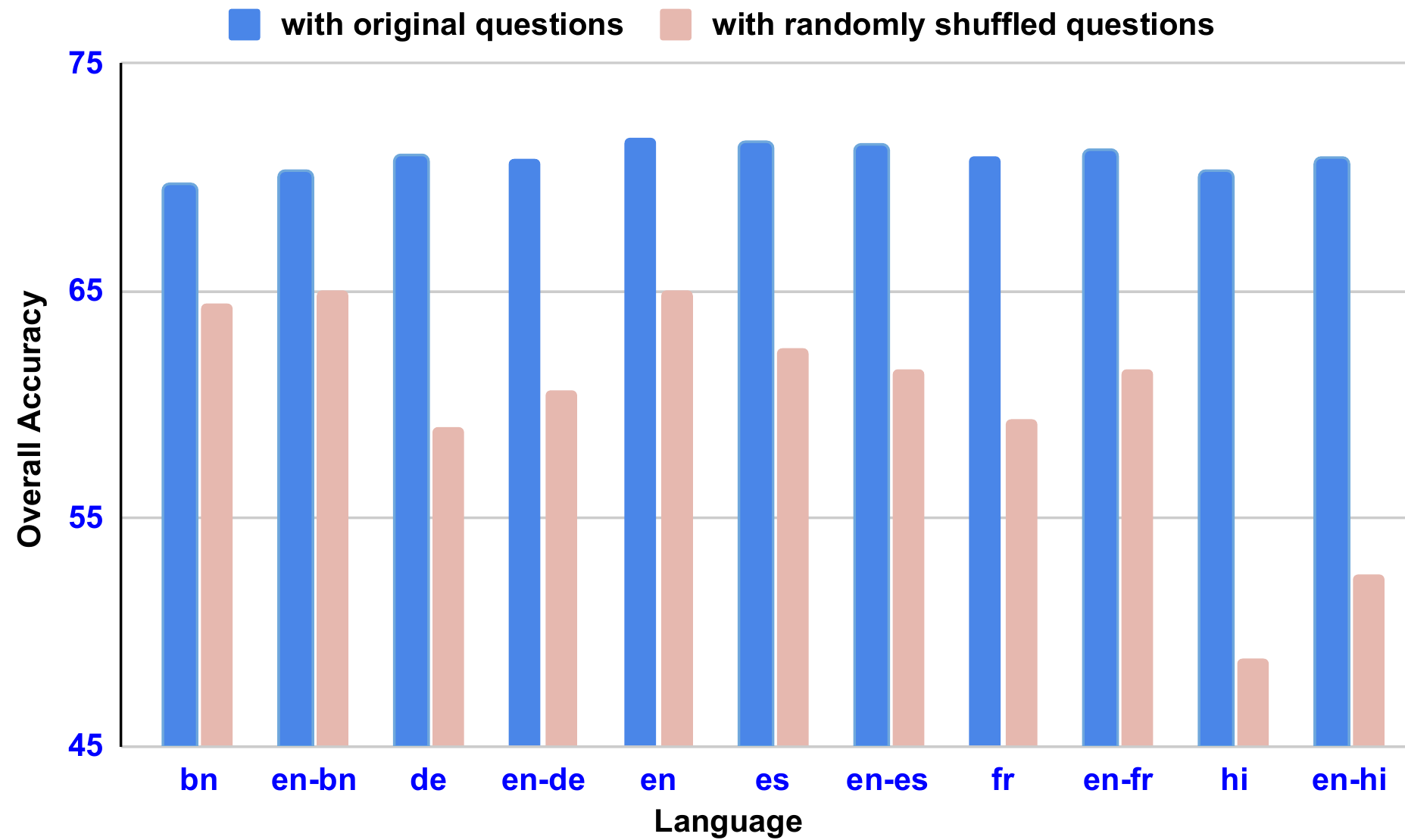}
    \caption{Performance comparison of the proposed model with shuffled questions and original questions}
    \label{fig:question-shuffle}
\end{figure}

\begin{figure}[]
    \centering
    \includegraphics[ width=\linewidth]{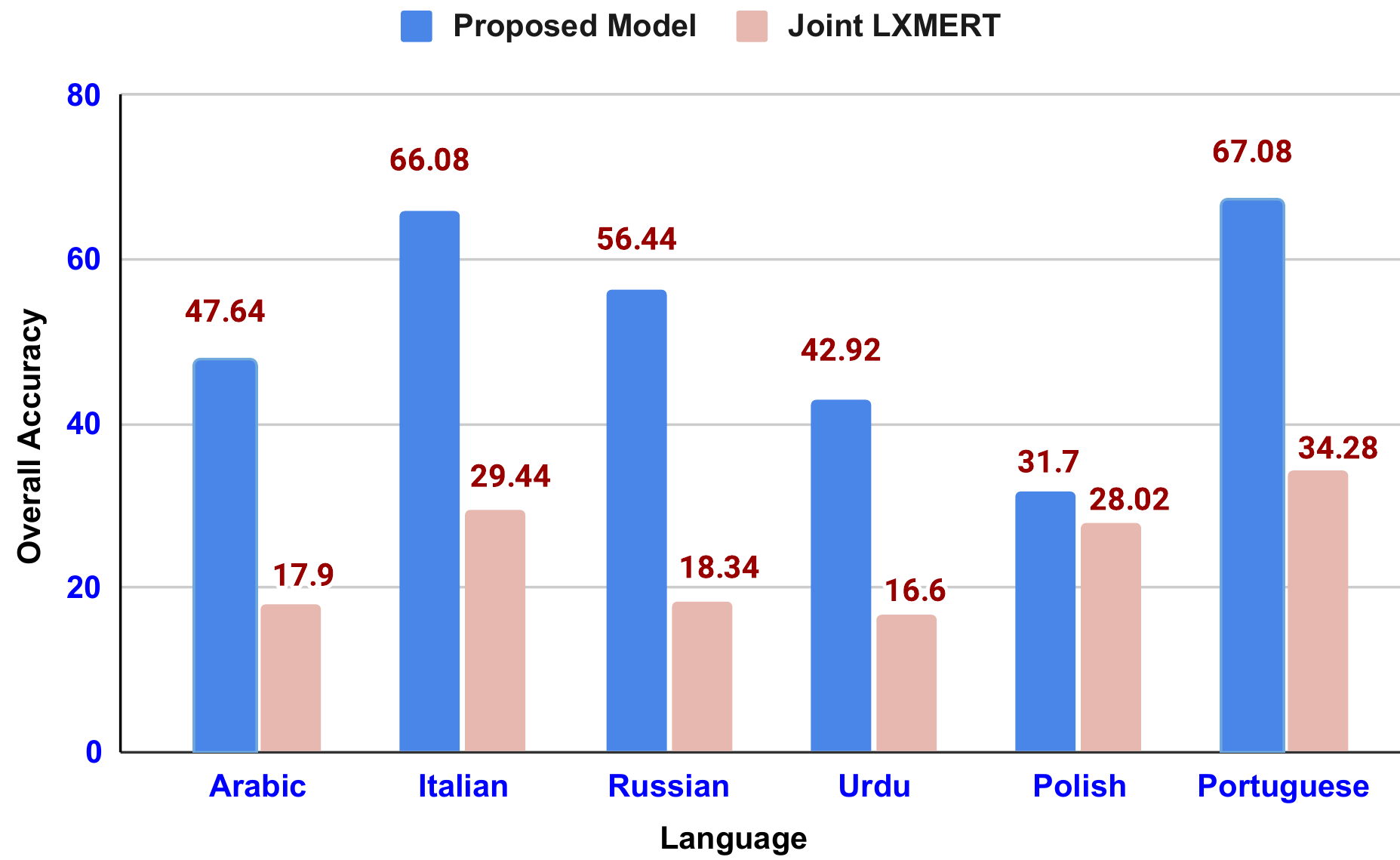}
    \caption{Zero-shot performance comparison of the proposed model on the different languages}
    \label{fig:zero-shot}
\end{figure}
% \begin{figure}[t]
%  \begin{minipage}{.22\textwidth}
%   \centering
%  {{\includegraphics[width=\linewidth]{May15/chart-2-main-paper.pdf} }}%
%  \end{minipage}
%     \quad
%     \begin{minipage}{.22\textwidth}
%   {{\includegraphics[width=\linewidth]{images/chart (1).pdf}}}%
%   \end{minipage}
% \captionof{figure}{Heatmap of the learned attention weight for various objects in the image from our proposed model. The proposed model is able to focus on the same object and correctly predict the answer irrespective of the language of the question.}
% \label{fig:question-object-alignment}
%  \end{figure}
%  \begin{figure}[]
%   \subfloat[\centering \textbf{Joint LXMERT}]{{\includegraphics[height=4cm,width=0.46\linewidth]{May15/chart.png}}}%
%     \quad
%     \subfloat[\centering \textbf{Proposed Approach}]{{\includegraphics[width=0.46\linewidth]{May15/zero.pdf} }}%
% \captionof{figure}{t-SNE visualization for MCM questions in all eleven language setups. For proposed approach (b), we observe that the question representations of the same questions (shown in the same color) in different languages are very close in vector space unlike the Joint LXMERT model (a).}
% \label{question-reps}
%  \end{figure}

   \begin{figure*}[h]
      \centering
      \includegraphics[width=0.95\linewidth]{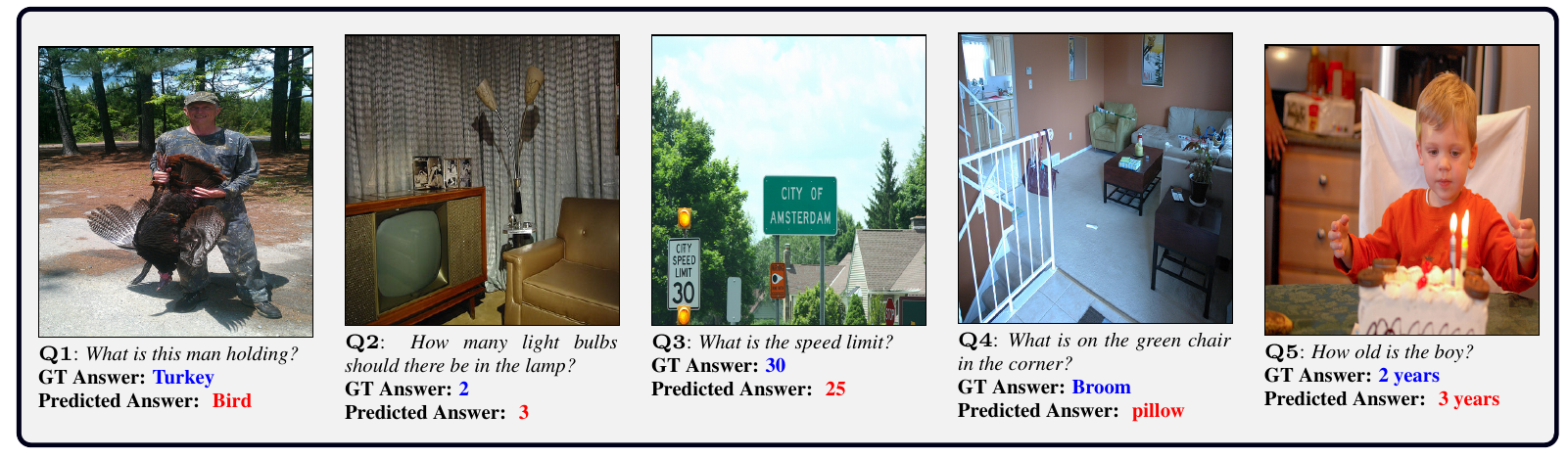}
      \caption{Examples from the various type of errors committed by our proposed model}
      \label{fig:vqa-error-analysis}
  \end{figure*}
 
\subsection{Discussion and Analysis} \label{sec:discussion}
\paragraph{Behavior Analysis:} 
We analyze the behavior of our proposed VQA model along the following dimensions:\\
\textbf{(a) Question Understanding}:
% Motivated from \citet{agrawal2016analyzing}, we investigated how much percent of the input question is used to make the decision. 
% Motivated from \citet{agrawal2016analyzing}, we investigated how the model responds to the partial questions in multilingual and code-mixed setups. 
% Motivated from \citet{agrawal2016analyzing}, we evaluate the behaviour of the model towards partial questions to asses their sen  of the model to the question by investigating how the model responds to the partial questions in multilingual and code-mixed setups.
Motivated from \citet{agrawal2016analyzing}, we analyze the performance of the model as a function of partial question length to establish the fact that the proposed model is more sensitive to MCM questions as compared to other pre-trained models.
% To examine this, we incrementally fed the whole questions to the LXMERT (monolingual), Joint-LXMERT, and proposed model (multilingual). The results show that our model reasons over the whole question to predict the answer. In contrast, other models look at only the first few words ($20$-$60\%$) of the questions to make the answer predictions.
To examine this, we fed the LXMERT (monolingual), Joint-LXMERT (multilingual), and the proposed model (multilingual) with partial questions in the range of $20$ to $100$\% in an incremental manner. We observe (\textit{c.f.} bar chart in Fig. \ref{fig:question-understanding-incomplete-main}) that our proposed model does not jump to quick conclusions by looking at partial questions as the overall accuracy is comparatively low for the proposed model for the incomplete questions. However, with full questions, the accuracies are high for the proposed model indicating that the model is sensitive to questions in different languages.\\
\indent Furthermore, we also analyze what percentage of answers do not change when the partial questions are provided as input to the model. 
% \indent Furthermore, we also evaluated what percentage of answers do not change when we incrementally provide question words to the model. 
%are fed to the model incrementally. 
% We show the percentage of questions (Y-axis) for which answers remain the same (compared to entire question) as a function of partial question length (X-axis) in Fig \ref{fig:question-understanding-incomplete}. 
% We can observe from the line chart of Fig-\ref{fig:question-understanding-incomplete} that for around $50\%$ of the questions the LXMERT and Joint LXMERT model, seems to have converged on a predicted answer after listening to just half the question. 
We can observe from the line chart of Fig. \ref{fig:question-understanding-incomplete-main} that our proposed model is capable to change the answers when more question words are received as input to the model, unlike the LXMERT and Joint LXMERT model where the answers remain the same for around $50\%$ of the questions.
% In contrast, our proposed model is capable to change the answers when more question words are received as input to the network. 
Additionally, to assess the role of syntax and semantics of the multilingual input questions, we analyze the performance of the system by feeding the randomly shuffled questions in Fig \ref{fig:question-shuffle}. The results show that our model is capable to understand the question semantics.\\
%We also report the additional results to assess the understanding of the proposed model for the shuffled questions in Fig \ref{fig:question-shuffle}.\\
% To assess the role of syntax and semantics of the multilingual input questions, we analyze the performance of the system by feeding the randomly shuffled questions. We observe (Fig \ref{fig:question-shuffle}) the significant decrements in the overall accuracy across the languages. We analyze the maximum (minimum) of $21.46\%$ ($5.16\%$) absolute decrements in overall accuracy for the \textit{hi} (\textit{bn}) language
\textbf{(b) Alignment}: 
We also analyze the alignment of the learned MCM question representation from our MCM Question Encoder. Towards this, we project the question representation (\texttt{[CLS]}) of the same question asked in different MCM settings using the t-SNE visualization \cite{van2008visualizing} in Fig \ref{question-reps}. The plot shows that the question representations learned from the Joint LXMERT model are scattered in the vector space. In contrast, our proposed model learns the question representations which are very close in the vector space, indicating the capability of the model to learn the language-agnostic question representations, which help the model to correctly predict the answer of the MCM questions.

In addition, we also analyze the cross-modal alignment learned from our proposed model. Towards this, we plot the attention heatmap (\textit{c.f.} Fig \ref{fig:question-object-alignment-main}) from the cross-modal encoder (\crossatt{}). We analyze that our proposed model is able to effectively learn the language-agnostic cross-modal representation, where the key objects from the images are attended to predict the correct answer for MCM questions. We also show (in \textbf{Appendix}) that the cross-modal representation learned from the proposed model is tightly coupled with the image and question as the attention to objects get change when the different questions are asked from the same image. 
Overall this analysis confirms that our model is not myopic to images and MCM questions to predict the answers.
\\
% The similar analysis is also performed and reported () for the LXMERT and Joint LXMERT model.\\
\textbf{(c) Zero-shot Capability}: We also assess the zero-shot capability of our proposed model. Towards this, we perform the experiments on the six more languages, \textit{viz.} Arabic (\textit{ar}), Italian (\textit{it}), Russian (\textit{ru}), Urdu (\textit{ur}), Polish (\textit{pl}), and Portuguese (\textit{pt}). We evaluate the performance of our proposed model in zero-shot manner on the $500$ questions translated into the respective languages (using Google translation). We compare the performance (\textit{c.f.} Fig \ref{fig:zero-shot}) with the multilingual Joint LXMERT model. 
% The results shows that our model is well-aligned in comparison to the Joint LXMERT model, to learn . 
% The proposed model learned to generate the well-aligned question representation from M-BERT model, achieves the better overall accuracy compared to the Joint LXMERT model. 
The proposed model achieves better overall accuracy compared to the Joint LXMERT model.
This demonstrates the capability of our model on the unseen languages, which eventually confirms that the proposed distillation objectives have guided the student to learn the robust cross-modal representations.
% This shows our model capability to  learn the cross-modalscale in any other language without any training data.
Please see the \textbf{Appendix} for the detailed qualitative analysis.
\vspace{-0.5 em}
\paragraph{Error Analysis:}
% To better understand the shortcomings and limitations of our model, we also perform a thorough analysis of the errors encountered by our proposed model on the \mucovqa{} dataset. We sampled $200$ questions where our model made the mistakes for all the languages. 
We categorize the following major sources of errors by sampling $200$ incorrectly predicted answers:\\
% \begin{itemize}[noitemsep]
\textbf{(a)} \textbf{Answer Specificity and Ambiguity (E1)}: This type of error occurs when the objects in the image can be interpreted in multiple ways based on their visual surroundings.  In those cases, our model sometimes predicts the incorrect but semantically similar to the ground truth (GT) answer. For example, $\mathbf{Q1}$ in Fig.  \ref{fig:vqa-error-analysis}, the question is \textit{``What is the man holding"}. Our model predicts the `\textit{bird}' as the answer for all languages of the questions. However, the ground truth answer is `\textit{Turkey}' which is more specific and semantically similar.\\
\textbf{(b)} \textbf{Object Counting (E2)}: We observe that our proposed model sometime predicts the incorrect answer for the counting type questions. The example is shown as $\mathbf{Q2}$ of Fig  \ref{fig:vqa-error-analysis}.\\
\textbf{(c)} \textbf{Character Recognition (E3)}: 
This type of error occurs when the answer to the MCM questions can only be predicted by recognizing the characters from the images. The example is shown in Fig. \ref{fig:vqa-error-analysis} ($\mathbf{Q3}$), where the GT answer is `$30$' (speed limit) but the model predicts the incorrect answer `$25$' because it could not recognize the character written in the image.\\
\textbf{(d)} \textbf{Spatial Interpretation (E4)}: 
Such errors occur when the model could not correctly interpret the spatial information in the image. The example is shown in Fig. \ref{fig:vqa-error-analysis} ($\mathbf{Q4}$), where the model predicted the `\textit{pillow}' as the answer instead `\textit{broom}'.\\
\textbf{(e)} \textbf{Answer Reasoning (E5)}: 
This type of error occurs for the question, which requires understanding the causal relationship or in-depth reasoning to correctly predict the answer. We show the example (Fig. \ref{fig:vqa-error-analysis} ($\mathbf{Q5}$)), where to infer the age of the boy, the system has to establish the fact that \textit{number of candles on the cake can determine the age}. 
There are some other errors caused by parallel question alignment and translation of the questions. We found the error \textbf{E5} contributes to the maximum of $26.5\%$, \textbf{E1}: $23.5\%$, \textbf{E3}: $21\%$, \textbf{E2}: $16\%$, \textbf{E4}: $9\%$ and other types of error contributes to $4\%$ of the total errors.
\section{Conclusion} \label{sec:conclusion}
In this paper, we have proposed a unified framework for multilingual and code-mixed VQA by distilling the knowledge from the monolingual language-vision pre-trained LXMERT model. To fully utilize the rich information from the question, image, and cross-modal encoders, we devise effective distillation objectives to encourages the student model to learn from the teacher through a multi-layer distillation process. To train and evaluate the proposed approach, we have created a large-scale \mucovqa{} dataset supporting eleven different MCM settings. Extensive experiments over the \mucovqa{} and MCVQA datasets demonstrate the effectiveness of our proposed approach. 
% Our experimental results and in-depth analysis show that our single unified model is able to answer the questions from different languages and code-mixed settings. 
% Our model also shows the remarkable zero-shot capabilities in six different languages. 

\section*{Ethical Declaration}
All the datasets used in this paper are publicly available. The dataset used in this paper is used only for the purpose of academic research. There are no ethical concerns associated with the research carried out here. 
\section*{Acknowledgement}
Asif Ekbal acknowledges the Young Faculty Research Fellowship (YFRF), supported by Visvesvaraya PhD scheme for Electronics and IT, Ministry of Electronics and Information Technology (MeitY), Government of India, being implemented by Digital India Corporation (formerly Media Lab Asia).

\bibliography{anthology,custom}
\bibliographystyle{acl_natbib}

\appendix

 \section{Multilingual and Code-Mixed VQA Dataset}
% \label{multilingual-dataset}
\subsection{\textbf{\mucovqa{} Dataset Creation}}

We use the Indic-nlp-library\footnote{\url{https://github.com/anoopkunchukuttan/indic_nlp_library}}
to tokenize the quesstions of the Indic languages and 
Moses based tokenizer\footnote{\url{https://github.com/moses-smt/mosesdecoder}} for remaining languages. Following, we learn the alignment matrix using the fast alignment technique proposed in \citet{dyer2013simple}. The alignment helps to select the words or phrases to be mixed in the code-mixed question. Thereafter, we construct the aligned phrases between the English and foreign language questions. We extract the PoS, named entity (NE), and noun phrase (NP) from the English questions, and mix them in the proper places of the corresponding Hindi questions. More specifically, we start with the NEs of types `\textit{PER}', `\textit{LOC}', and `\textit{ORG}' in the English question, and replace the corresponding words in the foreign language questions with the detected NEs from the English question. Similarly, we replace the corresponding words in the foreign language questions with the detected NPs from the English question. Finally, we also follow the same for the PoS tags `\textit{Adjective}'.
We utilize the constructed phrase and alignment information to identify the appropriate places to insert English words in the foreign language questions. 
% We show the samples from the created \mucovqa{} dataset in Figure \ref{fig:sample-mucovqa}.

\begin{figure*}
    \centering
    \includegraphics[ width=\textwidth]{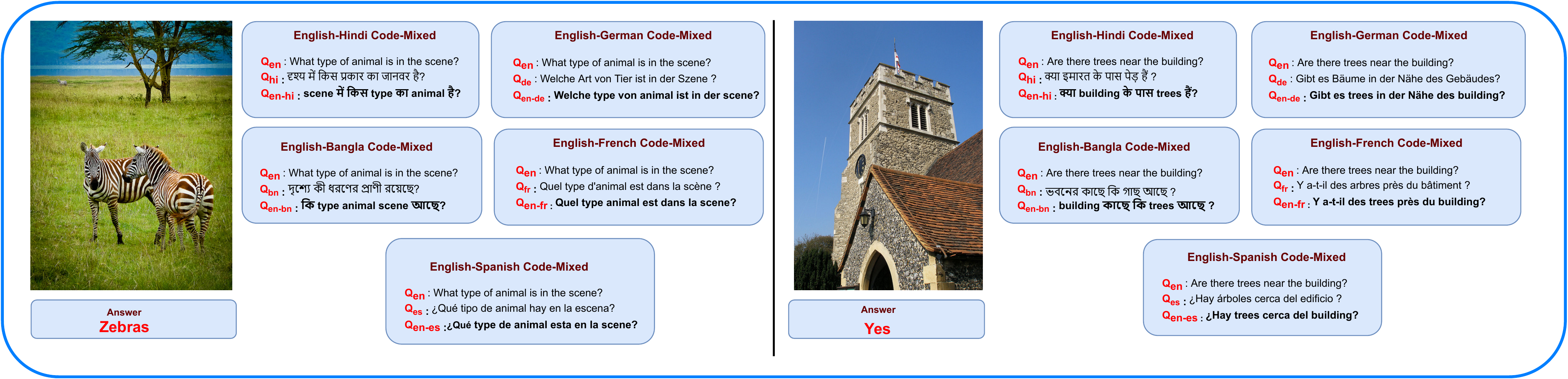}
    \caption{Sample questions (in multiple languages and code-mixed settings) with their corresponding images and answer from our \mucovqa{} dataset.}
    \label{fig:sample-mucovqa}
\end{figure*}

\subsection{\textbf{Analysis}}
We compute the complexity of the generated code-mixed questions using the Code-Mixing Index (CMI) \citep{gamback2014measuring}, Switch Point Fraction (SPF) \cite{pratapa2018language,gupta-etal-2020-semi} and Complexity Factor (CF) \citep{DBLP:journals/cys/GhoshGD17} for the entire code-mixed questions from \mucovqa{} dataset (Table \ref{tab:cmi-spf}) and aforementioned $500$ questions. These are the standard metrics used in the literature to indicate the level of language mixing in the code-mixed sentence. 
For the $500$ questions, the mean values of the individual score obtained from each human expert are shown in Table \ref{cm-data-statistics}. Our analysis shows that the code-mixed questions in \mucovqa{} dataset have similar CMI and SPF scores compared to the human formulated code-mixed questions. Similar observations are also made for the CF2 and CF3 metrics. The reported values in Table \ref{cm-data-statistics} also indicate that the automatically generated questions are slightly more complex (in terms of mixing the language) than the human-annotated code-mixed questions.

% \indent We also perform qualitative analysis by randomly selecting $5,200$ questions from our \texttt{MCVQA} dataset. A bilingual (En, Hi) expert was asked to manually create the code-mixed questions and translate the English questions into Hindi. We compute the BLEU \citep{papineni2002bleu}, ROUGE \citep{lin2004rouge} and Translation Error Rate (TER) \citep{snover2006study}
% on the human translated questions and the translations obtained from the Google Translate. We achieve high BLEU and Rouge scores (BLEU 3: $80.22$; ROUGE - L: $92.20$) and lower TER ($9.63$). 
% Similarly, we compute the complexity factor of automatically generated code-mixed questions, and found
% these to be highly comparable to the ground truths. 
% The detailed comparisons of automatically generated and manually annotated code-mixed questions are shown in Table \ref{cm-data-statistics}. 
% The scores are reported in Table \ref{cm-data-statistics}. %with the respective average scores on the automatically generated code-mixed questions. 
% (refer to the \textbf{Appendix} for more details).%Details of these scores with tabular comparisons of our MCVQA dataset with other VQA datasets in terms of multilinguality, code-mixing and data creation can be found in \textbf{Appendix}.

\begin{table}[h]
\centering
\resizebox{\linewidth}{!}{%
\begin{tabular}{c|c|c|c||c|c|c|c}
\hline
\textbf{Metrics} & \textbf{BLEU} & \textbf{ROUGE-L} & \textbf{TER} & \textbf{CMI} & \textbf{SPF} & \textbf{CF2} & \textbf{CF3} \\ \hline
\mucovqa{}  & $78.34$ & $91.13$   & $8.23$ & $33.42$  & $79.65$  & $13.14$  & $12.27$  \\ 
Human & NA & NA & NA & 33.23 & 80.21 & 13.43 & 12.59 \\ \hline
\end{tabular}%
}
\caption{Comparison of the generated code-mixed questions in terms of the level of code-mixing (CMI, SPF, CF2 and CF3) and quality of the generated code-mixed questions (BLEU, ROUGE-L and TER). Here, \textbf{NA}: Not applicable as the scores are computed against the human annotation itself.}
\label{cm-data-statistics}
\end{table}

 \begin{figure*}[!h]
    \centering
    \includegraphics[ width=\linewidth]{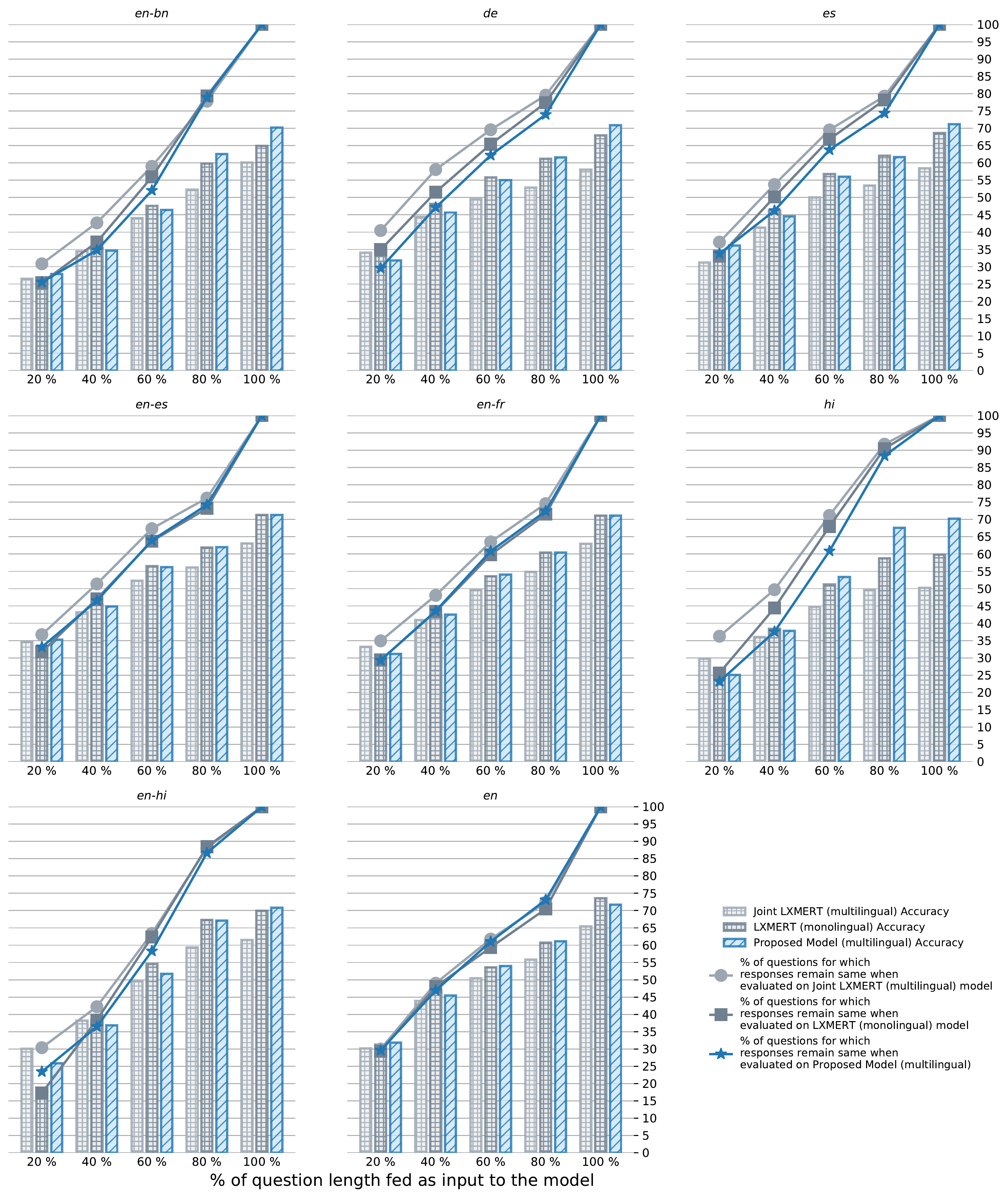}
    \caption{Performance comparison between the state-of-the-arts (LXMERT and Joint LXMERT) models and proposed model for question understanding by varying the partial question as input to the model.}
    \label{fig:question-understanding-incomplete}
\end{figure*}

% \usepackage{graphicx}
% \usepackage{graphicx}
\begin{table*}[!h]
\resizebox{\textwidth}{!}{%
\begin{tabular}{ccccccccc}
\hline
\textbf{\begin{tabular}[c]{@{}c@{}}Language\\ Pairs\end{tabular}} & \textbf{\begin{tabular}[c]{@{}c@{}}\#Code-Mixed \\ Question: Train\end{tabular}} & \textbf{\begin{tabular}[c]{@{}c@{}}\% of \\ Code-Mixed\end{tabular}} & \textbf{SPF} & \textbf{CMI} & \textbf{\begin{tabular}[c]{@{}c@{}}\#Code-Mixed\\  Question: Test\end{tabular}} & \textbf{\begin{tabular}[c]{@{}c@{}}\% of \\ Code-Mixed\end{tabular}} & \textbf{SPF} & \textbf{CMI} \\ \hline \hline
\textit{\textbf{en-bn}} & 243,203 & 97.93 & 92.47 & 35.65 & 118,989 & 97.92 & 92.21 & 36.14 \\ 
\textit{\textbf{en-de}} & 242,854 & 97.79 & 81.22 & 33.96 & 118,895 & 97.85 & 81.46 & 34.05 \\ 
\textit{\textbf{en-es}} & 234,570 & 94.45 & 74.80 & 31.69 & 114,747 & 94.43 & 74.80 & 31.70 \\ 
\textit{\textbf{en-fr}} & 241,430 & 97.21 & 80.27 & 33.98 & 118,112 & 97.20 & 80.17 & 33.93 \\
\textit{\textbf{en-hi}} & 242,963 & 97.83 & 78.35 & 32.82 & 118,935 & 97.88 & 78.54 & 32.80 \\ \hline \hline
\end{tabular}%
}
\caption{Statistics of generated code-mixed questions and along with the training and test set distributions. We also show the complexity of the generated code-mixed sentence in terms of
SPF and CMI}
\label{tab:cmi-spf}
\end{table*}

\begin{figure*}[h]
 \begin{minipage}{.1\textwidth}
  \centering
 {{\includegraphics[width=\linewidth]{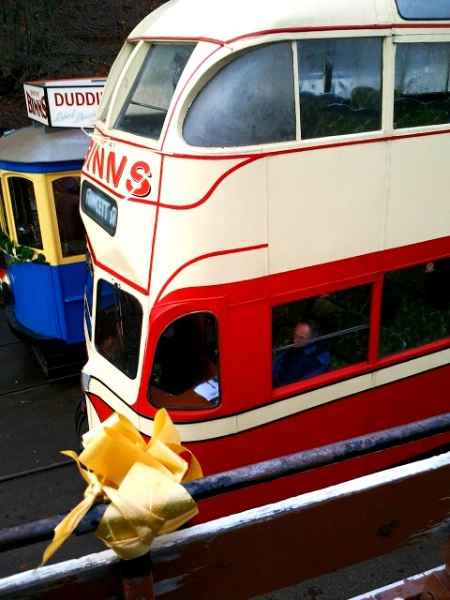} }}%
 \end{minipage}
    \quad
    \begin{minipage}{.9\textwidth}
  {{\includegraphics[width=\linewidth]{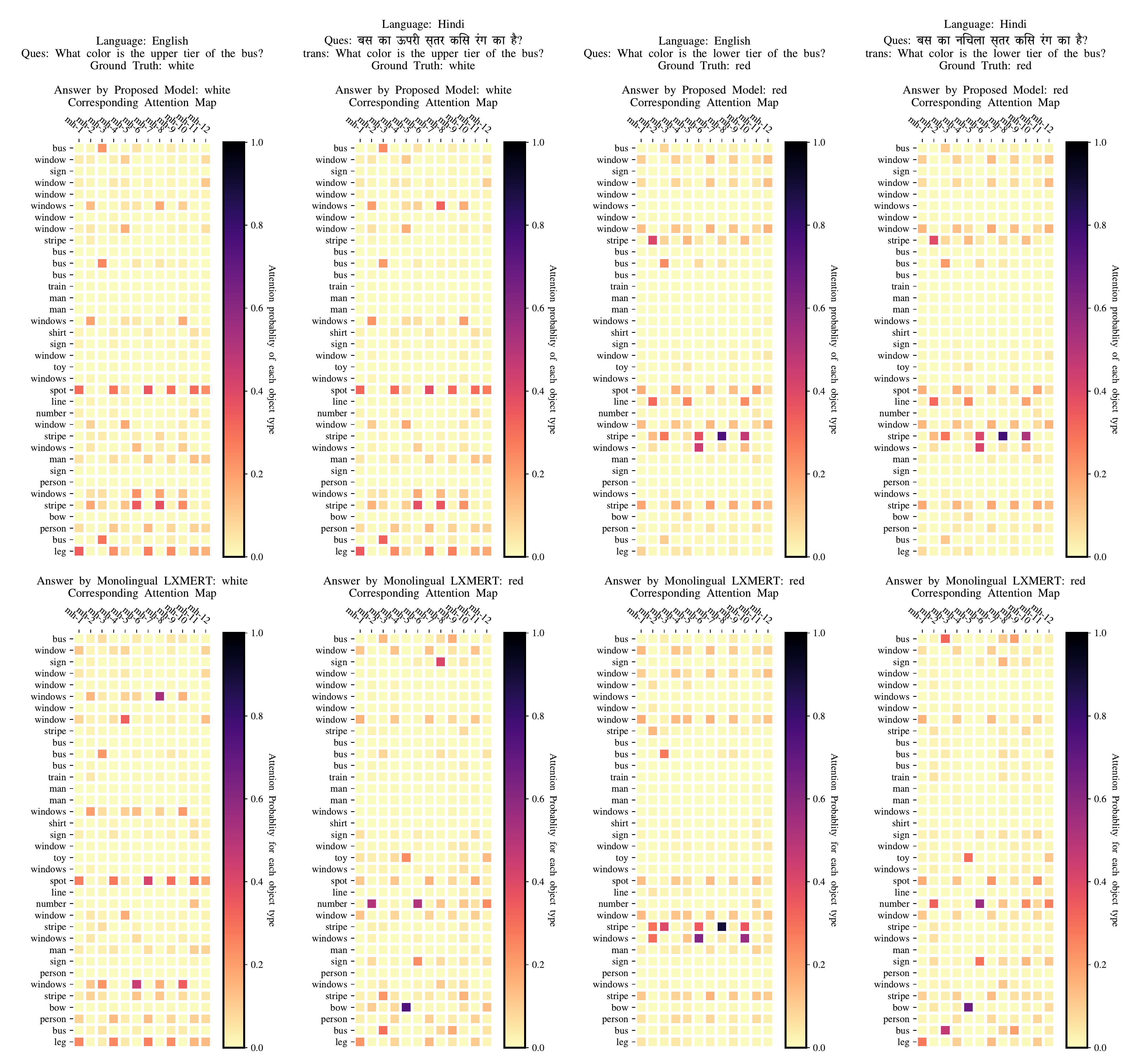}}}%
  \end{minipage}
\captionof{figure}{Heatmap of the learned attention weight for various objects in the image from our proposed model (Top) and LXMERT (Bottom). The proposed model is able to attend to the correct objects (the one attended by LXMERT when the English question is passed) in a language-agnostic way and hence predict the correct answer for MCM questions. However, the LXMERT monolingual model attends to the same objects and focuses only on the image giving same answers irrespective of the question. This shows the efficiency and robustness of the proposed model as it is sensitive to the question and maintains similar behavior across the languages.}
\label{fig:question-object-alignment}
 \end{figure*}

 \begin{figure*}[h]
      \centering
      \includegraphics[width=\linewidth]{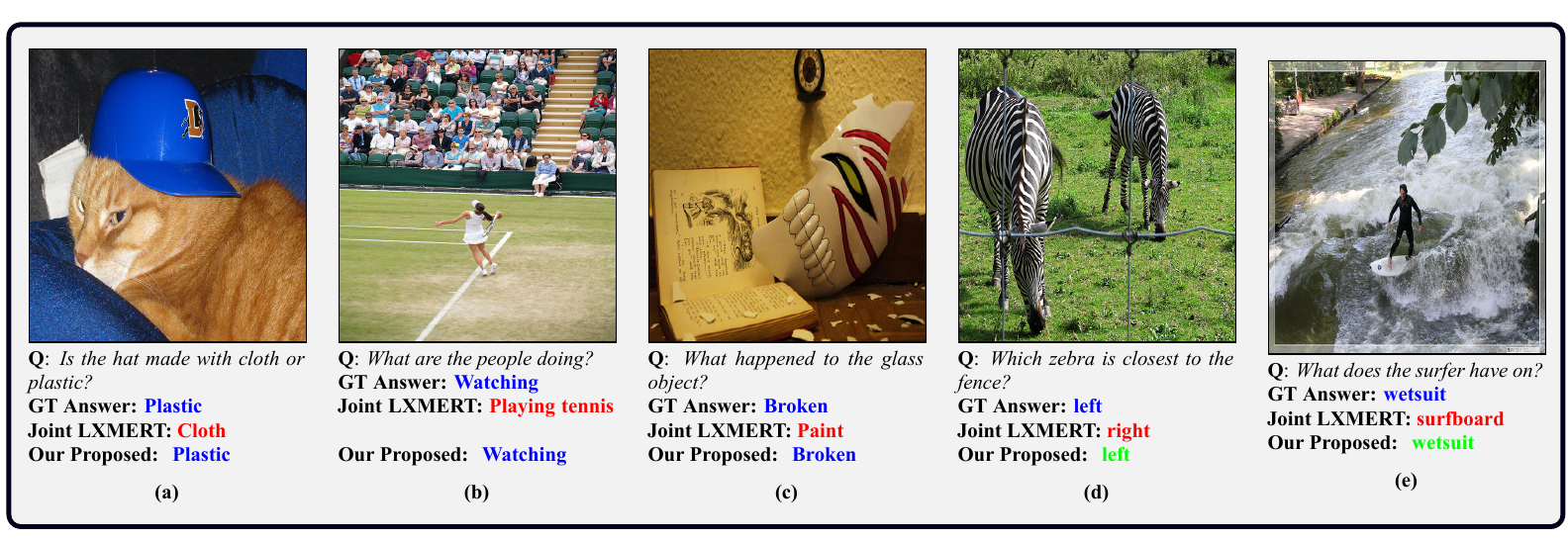}
      \caption{Sample questions where our proposed model perform better and correctly predict the answer compare to the multilingual Joint LXMERT model.}
      \label{fig:vqa-analysis}
  \end{figure*} 

\begin{table*}[!h]
\resizebox{\linewidth}{!}{%
\begin{tabular}{lcccccccccccc}
\hline
\textbf{Language}                    & \textbf{bn}    & \textbf{en-bn} & \textbf{de}    & \textbf{en-de} & \textbf{es }   & \textbf{en-es} & \textbf{fr}    & \textbf{en-fr} & \textbf{hi}    & \textbf{en-hi} & \textbf{en}    & \textbf{average} \\ \hline
\textbf{Validation Accuracy} & 72.98 & 73.51 & 74.08 & 73.88 & 74.29 & 74.52 & 74.25 & 74.43 & 73.42 & 74.13 & 74.86 & 74.03  \\
\hline
\end{tabular}
}
\caption{Performance of our proposed model on \mucovqa{ } validation dataset of different languages.}
\end{table*}

\section{\textbf{Teacher Network}}
\label{lxmert-background-sup}
% \paragraph{Teacher Network:}
Learning Cross-Modality Encoder Representations from Transformers (LXMERT) \cite{tan-bansal-2019-lxmert} is a pre-trained language model to learn the language-vision representation. It is built with the self-attention and cross-attention layers. The LXMERT model is pre-trained with a large amount of image-and-sentence pairs from VQA v2.0 \citep{goyal2017making}, GQA \citep{Hudson_2019_CVPR}, and VG-QA \citep{zhu2016cvpr} datasets. It is pre-trained on different tasks, such as \textit{masked object prediction}, \textit{masked language modeling}, \textit{visual question answering}, and \textit{cross-modality matching}.

Given a text and an image as inputs, LXMERT learns the language, image, and cross-modality (language-image) representations from the inputs. The language embedding is created using the word and position embeddings followed by applying the layer normalization operation on the embeddings. The language encoder, which is composed of Transformer encoders takes the language embedding as input and generates the language representation. The image embedding is generated using the features of the detected objects from the image. Each detected object in the image is represented by its position and region-of-interest (RoI) features. The final image embedding is computed by averaging the revised position and RoI features using the layer normalization operation on the respective feature. The image embedding is passed into the image encoder, which is another transformer encoder. The cross-modality encoders are the stack of multiple encoder layers. Each encoder layer consists of two self-attention sub-layers, one bi-directional cross-attention sublayer, and two feed-forward sub-layers. The bi-directional cross-attention sub-layer contains one sub-layer from language to image and another from image to language.
\section{Additional Implementation Details}
To update the model parameters, we use the Adam \cite{adam} optimization algorithm with the learning rate of $1e-5$. We obtain the optimal hyper-parameter values based on the performance of the model on the validation set of \mucovqa{} dataset. We use a cosine annealing learning rate \cite{sgdr} decay schedule, where the learning rate decreases linearly from the initial rate set in the optimizer to $0$. To avoid the gradient explosion issue, the gradient norm was clipped within $6$. For doing the baseline experiments, we follow the official source code and train the model on the \mucovqa{} dataset. All the experiments are performed on a single GeForce GTX 1080 Ti GPU having GPU memory of $11$GB. The average runtime (each epoch) for the proposed approach is $2.5$ hrs. 
% The proposed student model has $291$ millions parameters out of which $132$ millions are trainable parameters and teacher LXMERT model has $241$ millions trainable parameters.

\end{document}